%% file: main.tex
\documentclass{article} 
\usepackage{iclr2026_conference,times}

\input{math_commands.tex}

\usepackage{hyperref}       
\usepackage{url}            
\usepackage{amsmath, amsfonts}
\usepackage[table]{xcolor}
\usepackage{booktabs}       
\definecolor{bestgreen}{HTML}{E7FBFF}
\definecolor{secondbluebase}{HTML}{E7FBFF}
\colorlet{secondblue}{secondbluebase!40!white}
\newcommand{\second}[1]{\cellcolor{secondblue}#1}
\newcommand{\best}[1]{\cellcolor{bestgreen}{#1}}
\usepackage{subcaption}     
\usepackage{amsfonts}       
\usepackage{nicefrac}       
\usepackage{microtype}      
\usepackage{multirow}
\usepackage{amsmath}
\usepackage{mathtools} 
\usepackage{xcolor}         
\usepackage{makecell}       
\usepackage{caption}
\usepackage{enumitem}       
\usepackage{graphicx}
\usepackage{subcaption}
\setlist[enumerate]{leftmargin=2em, labelsep=0.5em, label=\arabic*.}

\setlist[itemize]{leftmargin=2em, labelsep=0.5em, itemsep=0.2ex, topsep=0.6ex}

\title{
STAGE: Stable and Generalizable GRPO for Autoregressive Image Generation
}

\author{%
  Xiaoxiao Ma$^{1,2}$\,\space\space\space
  Haibo Qiu$^{2}$\thanks{Project lead; $^{\dagger}$ Corresponding author.}\,\space\space\space
  Guohui Zhang$^{1}$\,\space\space\space
  Zhixiong Zeng$^{2}$\space\space\space
  Siqi Yang$^{2}$\space\space\space
  \\
  \vspace{0.1cm}
  \textbf{Lin Ma$^{2}$\space\space\space Feng Zhao$^{1,\dagger}$}
  \\
  $^{1}$\normalfont{University of Science and Technology of China}\quad
  $^{2}$\normalfont{Meituan}
  \\
  \vspace{0.08cm}
  \parbox{\textwidth}{%
    \centering
    \begin{tabular}{@{}p{0.55\linewidth} p{0.45\linewidth}@{}}
      \tt\small \{xiao\_xiao,guohuizhang\}@mail.ustc.edu.cn & \tt\small haibo-qiu@outlook.com \\
      \tt\small zengzhixiong@meituan.com & \tt\small siqi.yang@uq.net.au \\
      \tt\small forest.linma@gmail.com & \tt\small fzhao956@ustc.edu.cn \\
    \end{tabular}%
  }%
}

\iclrfinalcopy 
\begin{document}

\maketitle
\input{sec/abstract}
\input{sec/introduction}
\input{sec/related_works}
\input{sec/method}
\input{sec/experiments}
\input{sec/conclusion}

\clearpage
\bibliography{iclr2026_conference}
\bibliographystyle{iclr2026_conference}

\appendix
\input{sec/supp/add_method_description}
\input{sec/supp/detailed_exp_setting}
\input{sec/supp/add_metric}
\input{sec/supp/relation_entropy_image}
\input{sec/supp/analysis_diverse}
\input{sec/supp/limitations}
\clearpage
\input{sec/supp/add_vis_comp}

\end{document}

%% file: math_commands.tex
\usepackage{amsmath,amsfonts,bm}

\def\eqref#1{equation~\ref{#1}}

\def\1{\bm{1}}

\DeclareMathAlphabet{\mathsfit}{\encodingdefault}{\sfdefault}{m}{sl}
\SetMathAlphabet{\mathsfit}{bold}{\encodingdefault}{\sfdefault}{bx}{n}



%% file: sec/abstract.tex
\begin{abstract}
\label{abstract}
Reinforcement learning has recently been explored to improve text-to-image generation, yet applying existing GRPO algorithms to autoregressive (AR) image models remains challenging. The instability of the training process easily disrupts the pretrained model capability during long runs, resulting in marginal gains, degraded image quality, and poor generalization. In this work, we revisit GRPO for AR image generation and identify two key issues: contradictory gradients from unnecessary tokens and unstable policy entropy dynamics. To address these, we introduce STAGE, a stable and generalizable framework that leverages two targeted solutions: 1) Advantage/KL reweighting. Similarity-aware reweighting to alleviate conflicting updates; and 2) Entropy reward. An entropy-based reward corresponding to reference model to stabilize learning. With the help of alleviating conflicts between tokens and an entropy reward for stabilizing training, we reduce disruption of the pretrained distribution and mitigate reward hacking, which in turn improves generalization and transfer better to other benchmarks.
Experiments across multiple benchmarks show that STAGE consistently improves visual quality, stability, and cross-task generalization compared to baseline GRPO. Code is available at \textcolor{cyan}{\url{https://github.com/krennic999/STAGE}}.
\end{abstract}

%% file: sec/introduction.tex
\section{Introduction}
\label{introduction}
Reinforcement learning (RL) for large language models (LLMs) has markedly improved performance on reasoning-intensive tasks such as mathematics and code generation. In particular, Group Relative Policy Optimization (GRPO)~\citep{shao2024deepseekmath} eliminates the value model in PPO~\citep{schulman2017ppo}, resulting in a simpler and more efficient training paradigm via group-relative advantages. Recent advances further strengthen this with importance weighting~\citep{zheng2025gspo,zhao2025gmpo} and entropy regularization~\citep{cui2025entropyrl,wang2025_2080rule}, establishing RL as a powerful paradigm for performance gains and human-preference alignment.

Corresponding RL technique have also been explored for visual generation. For continuous-representation models,~\citet{liu2025flow_grpo,xue2025dance_grpo} investigated GRPO for flow models~\citep{esser2024sdv3.5,flux2024}. For discrete autoregressive (AR) approaches, T2I-R1~\citep{jiang2025t2i_r1} pairs semantic reasoning with RL, AR-GRPO~\citep{yuan2025ar_grpo} directly applies GRPO for AR, Janus-focusdiff~\citep{pan2025focusdiff} integrates RL into fine-tuning, and SimpleAR~\citep{wang2025simplear} applies RL in unified vision–language models to improve image quality.
RL has enhanced current generative foundation models and offers a promising direction for visual reasoning.

Nevertheless, current GRPO adaptations for AR image generation largely follow LLM practices and do not fully account for the characteristics of visual tokens.
In GRPO, at each RL iteration the policy generates a group of samples per prompt, a reward (e.g., HPS~\citep{wu2023hpsv2} or GenEval~\citep{ghosh2023geneval}) scores each sequence, and scores are propagated token-wise to update the policy.
Yet, \textbf{visual tokens differ fundamentally from text}: 1) Although discrete, they represent continuous semantics align with low-level patterns in the decoded image. 2) Rollouts from the same prompt often share highly similar content, especially background regions (see Fig.~\ref{fig_motivation} (b)). However, when GRPO enforces divergent updates, semantically similar regions across rollouts may be assigned opposite rewards, introducing noisy and contradictory gradients.
3) Due to the discrete and sequential nature of AR generation, AR models are highly sensitive to small distribution shifts. Especially under repeated RL optimization, the model struggles to maintain a stable distribution, leading to reward hacking or degraded outputs and poor generalization (see Fig.~\ref{fig_motivation}(a)). KL regularization can partially alleviate this issue, yet some risk of performance degradation remains.

To address these challenges, we propose STAGE, which augments GRPO paradigm with two targeted improvements: 
1) similarity-aware advantage/KL reweighting to improve training efficiency, and 2) an entropy-based regularizer to stabilize learning.
Specifically: 1) Token-wise advantage leverages similarities among token embeddings within rollouts of the same prompt to dynamically adjust per-token advantages, reducing updates on redundant similar background tokens, mitigating conflicting gradients between positive and negative samples, and better preserving the original model capability. 
A similarity-aware KL schedule further suppresses unnecessary updates in irrelevant regions.
2) Entropy-based regularization calculates the entropy gap between current and reference policies and incorporates it as an auxiliary reward, further discouraging abrupt entropy drops and stabilizing policy updates.
Together, these mechanisms produce a more stable and efficient RL process (see Fig.~\ref{fig_ours_geneval}), mitigating reward hacking and improving generalization across image-quality metrics.

Extensive experiments on GenEval, T2I-CompBench and HPS show that our method outperforms baseline GRPO in stability and generalization.
With proposed approach, Janus-Pro’s GenEval score rises from 0.78 to 0.89, significantly surpassing most current diffusion and AR models. Training under HPS, OCR, and other rewards further demonstrates improvements in image quality and text rendering.
Notably, our method maintains stable entropy during RL while improving GenEval performance, achieving a favorable balance between visual detail, structural consistency, and prompt adherence. Compared with the baseline, it shows stronger generalization to prompts outside the training distribution.
We summarize our contributions as follows:
\begin{enumerate}
    \item Motivated by the challenge that GRPO for autoregressive image generation often suffers from unstable training and poor generalization, we propose STAGE, which addresses contradictory gradients and unstable entropy during training, improving efficiency while mitigating instability.
    \item Specifically, to handle contradictory gradients, we exploit similarities among multiple rollouts of the same prompt to avoid updates in regions shared by positive and negative samples, providing a smoother and more efficient RL process. An additional entropy-based reward that regularizes the current and reference policies further stabilizes training.
    \item Experiments across diverse rewards and benchmarks show that proposed method stabilizes training and improves detail and structural consistency in generated images. It also shows better generalization than baseline to evaluation metrics and prompts out of training distribution.
\end{enumerate}

%% file: sec/related_works.tex
\section{Related Works}
\label{related_works}

\subsection{Autoregressive Image Generation}
For AR image generation, images are first quantized into discrete tokens~\citep{esser2021vqgan,yu2021vit_vqgan} and then generated with Transformers in raster order~\citep{ding2021cogview,ge2023seed,ramesh2021dalle,yu2022parti,he2024mars,wang2024emu3}.
Recent efforts have scaled this paradigm with larger models and stronger conditioning. LlamaGen~\citep{sun2024llamagen} provides class and text-conditioned baselines; while~\cite{liu2024lumina_mgpt} and~\cite{chern2024anole} fine-tune Chameleon~\citep{chameleonteam2025chameleon} for improved text-conditioned generation. 

Recent work has explored unified vision-language generation, producing images and text within a single transformer~\citep{wu2024janus,chen2025janus_pro,jiao2025unitoken,unitok,zhang2025v2flow,qu2024tokenflow}, more powerful tokenizers~\citep{lee2022rqvae,yu2023lfq}, and strategies for parallel multi-token generation or token compression~\citep{tian2024var,ma2024star,yu2024titok,liu2025detailflow}. Training AR models typically involves multiple stages to better utilize limited high-quality data, and prior studies~\citep{sun2024llamagen,chen2025janus_pro} emphasize that carefully designed curricula are crucial for strong generative performance.

\subsection{Reinforcement Learning}
Reinforcement learning (RL) has been widely adopted in large language models (LLMs) to improve reasoning and alignment~\citep{deepseekai2025deepseekr1}. GRPO~\citep{shao2024deepseekmath} simplifies policy optimization by removing the value model in PPO~\citep{schulman2017ppo} and achieves strong empirical gains using a relative group-based objective. Subsequent refinements, including importance sampling~\citep{zheng2025gspo,zhao2025gmpo}, gradient clipping~\citep{yu2025dapo} and entropy-based regularization~\citep{cui2025entropyrl,wang2025_2080rule}, which further stabilize training.

In visual generation, RL has been used to enhance fidelity and controllability. Flow-based approaches~\citep{liu2025flow_grpo,xue2025dance_grpo} apply GRPO to align continuous generative processes~\citep{flux2024,esser2024sdv3.5} with human preferences~\citep{kirstain2023pickscore,wu2023hpsv2} or prompt alignment~\citep{huang2023t2i_compbench,ghosh2023geneval}. For AR models, RL is applied differently: T2I-R1~\citep{jiang2025t2i_r1} uses semantic reasoning to improve text-to-image alignment, AR-GRPO~\citep{yuan2025ar_grpo} provides a direct AR+GRPO baseline, and SimpleAR~\citep{wang2025simplear} integrates RL into unified model to improve quality. Despite these advances, current GRPO for AR image generation still suffers from inefficiency, unstable training and sub-optimal generalization.

%% file: sec/method.tex
\section{Method}
\label{sec_method}
\subsection{Preliminaries}
\label{sec_preliminary}

\noindent\textbf{Autoregressive image generation.}
In a standard autoregressive (AR) generation pipeline, an image $I \in \mathbb{R}^{H \times W \times 3}$ is discretized into a sequence of tokens $(x_1, x_2, \ldots, x_{h \times w})$, where each token $x_t\in[V]$ corresponds to an index in a learned codebook of size $V$ (e.g., from a VQ-VAE tokenizer). Given corresponding text tokens $c$, the transformer is trained to model the joint distribution of image tokens in a flattened sequence, where causal attention restricts each position to attend only to preceding tokens. During generation, tokens are produced in a raster-scan manner (from top-left to bottom-right). At step $t$, the model predicts a categorical distribution over the vocabulary conditioned on all previously generated tokens $x_{1:t-1}$. The overall generation process can thus be factorized as:
\begin{equation}
    p(x_{1:h\times w}) = \prod_{t=1}^{h\times w} p(x_t \mid x_{1:t-1}; c),
\end{equation}
where $p(x_t \mid x_{1:t-1};c)$ represents the conditional distribution of the $t$-th token on previous $t-1$ image and text tokens. This sequential factorization models long-range dependencies across visual tokens but also makes generation sensitive to distributional shifts accumulated along the sequence.

\noindent\textbf{Group relative policy optimization (GRPO).}
For image generation, GRPO improves downstream metrics via an iterative \emph{generate–evaluate–update} process. 
Concretely, the policy model $\pi_\theta$, parameterized as an AR transformer, produces $G$ diverse image outputs $\{o_1,\dots,o_G\}$ conditioned on $c$. 
Each output is then scored by a reward function $\mathcal{R}(\mathbf{x},c)$ to obtain rewards $\{R_1,\dots,R_G\}$. 
Advantages for each token $t$ in sample $i$ are computed by normalizing rewards within the group:
\begin{equation}
\hat{A}_{i,t} = \frac{R_i - \text{mean}(\{R_j\}_{j=1}^G)}{\text{std}(\{R_j\}_{j=1}^G)}.
\label{eq_preliminary_advantage}
\end{equation}
Subsequently, using importance sampling, policy $\pi_\theta$ is updated by maximizing following objective:
\begin{equation}
\mathcal{J}_\text{GRPO}(\theta) =
\mathbb{E}\Bigg[ \frac{1}{G}\sum_{i=1}^G \frac{1}{|o_i|}\sum_{t=1}^{|o_i|}
\min\!\big(r_{i,t}(\theta)\hat{A}_{i,t},\,
\text{clip}(r_{i,t}(\theta),1-\varepsilon,1+\varepsilon)\hat{A}_{i,t}\big)
- \beta D_\text{KL}(\pi_\theta\|\pi_\text{ref}) \Bigg],
\label{eq_grpo_opt_objective}
\end{equation}
where $\pi_\text{ref}$ denotes the reference policy for regularization and prevent from distributional drift.  
The importance ratio $r_{i,t}(\theta)$ is defined as:
\begin{equation}
r_{i,t}(\theta) = \tfrac{\pi_\theta(o_{i,t}\mid o_{i,<t};c)}{\pi_{\theta_\text{old}}(o_{i,t}\mid o_{i,<t};c)},
\label{eq_important_ratio}
\end{equation}
which measures the relative likelihood of token $o_{i,t}$ under current policy $\pi_\theta$ and old policy $\pi_{\theta_\text{old}}$.
For brevity, we omit the mapping from token sequences to rendered images. 
Here, $o_{i,t}$ denotes the $t$-th token of $i$-th sample, and $o_i$ is $i$-th generated image or corresponding full token sequence.

\subsection{Advantage \& KL Reweight}
\label{sec_method_advantage_reweight}
\noindent\textbf{Instability during AR training.}
During GRPO training of AR image generation models, we observe persistent instability. For example, under the GenEval reward, late-stage RL often produces structurally degraded images and plateaued evaluation scores, particularly with large learning rates, many iterations, or low KL loss weight. We attribute this to disruption of the pretrained model distribution (see Fig.~\ref{fig_motivation}(a)). As training progresses, the model’s prior knowledge of specific concepts, such as bicycles, is gradually degraded, resulting in deteriorated generated structures.

Addressing this issue requires better preservation of the pretrained distribution during RL training.
We attribute this to unstable training and noisy or conflicting gradients in RL objective, which obscure the optimization direction and cause distributional drift, ultimately degrading training outcomes.

\noindent\textbf{The conflicts in generated images.}
As discussed in Sec.~\ref{sec_preliminary}, at each training step the policy $\pi_{\theta}$ generates multiple outputs $\{o_i\}$ conditioned on text $c$.  
For AR visual generation, outputs from the same prompt often share highly similar regions, differing only in fine details (see Fig.~\ref{fig_motivation} (b)).

Unlike text tokens in LLMs, visual tokens exhibit strong local similarity. Although different images may receive different advantage values, many regions across outputs are nearly identical (even if token indices differ, their VQ embeddings are close), which results in conflicting gradients when advantages have opposite signs.

To quantify this, we compute token-level cosine similarity between VQ embeddings. For images $o_i$ and $o_j$, with embeddings $q_{i,t}, q_{j,t} \in \mathbb{R}^C$ at position $t \in \{1,\dots,h \times w\}$, we define and rewrite $\text{cos}(q_{i,t}, q_{j,t})$ as $\text{cos}(i,j,t)$, the cosine similarity is calculated as:
\begin{equation}
\text{cos}(i,j,t) = \frac{q_{i,t} \cdot q_{j,t}}{\|q_{i,t}\| \, \|q_{j,t}\|}.
\label{eq_cal_cos}
\end{equation}

As shown in Fig.~\ref{fig_motivation} (c), regions with similar content exhibit notably higher similarity scores. 
Even when cosine similarity is only slightly above zero, corresponding regions remain visually alike, while in most areas both cosine similarity and visual content across images are consistently high.
Motivated by this observation, we aim to discard redundant tokens or reduce their update during training.
\input{figs/motivation}

\noindent\textbf{\textit{Solution}: Advantage \& KL reweighting.}
Based on the previous observation, we leverage embedding similarity between samples with opposite advantage signs to identify and mitigate conflicting gradients.
For a group of images $\{o_1,\dots,o_G\}$ normalized advantages 
$\{\hat{A}_{i,t}\}_{i=1}^G$, For each sample $i$ at position $t$, we compute the cosine similarity \text{cos}(i,j,t) based on Eq.~\ref{eq_cal_cos}, and aggregate similarities only with tokens from samples $j$ whose advantages have opposite signs:
\begin{equation}
\text{Sim}(i,t) = 
\frac{\sum_{j=1}^G \mathbf{1}[\hat{A}_{i,t}\hat{A}_{j,t} \leq 0] \cdot \text{cos}(i,j,t)}
{\sum_{j=1}^G \mathbf{1}[\hat{A}_{i,t}\hat{A}_{j,t} \leq 0]}.
\label{eq_embed_sim}
\end{equation}
We then define a soft mask to down-weight highly similar tokens:
\begin{equation}
M_{i,t} = \mathrm{Norm}(1-\mathrm{Sim}(i,t)),
\end{equation}
where $\mathrm{Norm}(\cdot)$ scales and shifts $M_{i,t}$ from range $[-1,1]$ to $[0,1]$.
And the masked advantage is then
\begin{equation}
\tilde{A}_{i,t} = M_{i,t} \cdot \hat{A}_{i,t}.
\label{eq_embed_advantage_reweight}
\end{equation}

By replacing $\hat{A}_{i,t}$ in Eq.~\ref{eq_preliminary_advantage} with the weighted form $\tilde{A}_{i,t}$, we obtain the similarity-modulated advantage. 
Intuitively, tokens similar to opposite-advantage tokens are down-weighted to avoid contradict gradients, while less similar tokens are amplified for clearer optimization direction.

To further stabilize training, we dynamically reweight the per-token KL penalty based on embedding similarity. Specifically, we assign a per-token KL weight based on embedding distance between the positive and negative samples at the same position: when the embeddings are close (high model certainty), we apply a larger KL penalty, and vice versa.
Based on the similarity $\text{Sim}(i,t)$ from Eq.~\ref{eq_embed_sim}, the per-token KL weight is computed as $\beta'_{i,t} = (a + b \cdot \mathrm{clip}(\text{Sim}(i,t)+1)) \, \beta$, where $a = b = 0.5$.

\subsection{Entropy Reward for Stablized Training}
\label{sec_method_entropy_reward}
\noindent\textbf{Entropy \& generated images.} 
We first define entropy in the context of autoregressive generation.  
Given a policy $\pi_\theta$ and conditional distribution over token vocabulary at step $t$, entropy is defined as:
\begin{equation}
    \mathcal{H}_t = - \sum_{x \in V} \pi_\theta(x \mid x_{<t}; c) \log \pi_\theta(x \mid x_{<t}; c),
    \label{eq_cal_entropy_token}
\end{equation}
where $V$ denotes the vocabulary and $c$ the input condition. 
The overall entropy of a generated sequence $o=(o_1,\dots,o_T)$ is obtained by averaging across positions: 
\begin{equation}
    \mathcal{H}(o) = \frac{1}{T}\sum_{t=1}^T \mathcal{H}_t.
    \label{eq_cal_entropy}
\end{equation}

Entropy plays a crucial role in AR image generation. 
A low-entropy policy tends to produce highly deterministic outputs, possibly leading to a loss of diversity, 
while an excessively high-entropy policy encourages randomness that results in noisy or semantically inconsistent generations.

\input{figs/motivation_entropy}

To study this relationship, we perform a controlled study by varying sampling temperature $\tau$ of a fixed AR model during generation, which directly controls token-level entropy.
As shown in Fig.~\ref{fig_motivation_entropy}, decreasing $\tau$ leads to more deterministic samples and may lower image quality, while increasing $\tau$ enriches content at the cost of structural accuracy.
(See Appendix~\ref{supp_add_analysis_entropy} for additional analysis).
These results indicate that maintaining an appropriate entropy range is essential for balancing fidelity and diversity in generation, 
and motivate the introduction of entropy-aware reward during RL fine-tuning.

\noindent\textbf{Entropy collapse during RL.}
During RL training, the policy entropy of each generated sample $\{o_1, \dots, o_G\}$ 
is defined based on the probability distributions of all tokens obtained by feeding 
the generated sample into the policy $\pi_\theta$, and computed according to 
Eq.~\ref{eq_cal_entropy_token} and Eq.~\ref{eq_cal_entropy}.

Training with different rewards induce distinct policy entropy dynamics.
In particular, for VQA- or rule-based rewards (e.g., GenEval reward in Flow-GRPO~\citep{liu2025flow_grpo}), in some cases, qualitatively different images can receive the same reward. As shown in Fig.~\ref{fig_motivation_entropy}, as long as the specified object from the prompt appears in the image, the output is
judged as correct. This leads to confusion in the policy model during learning, potentially causing different entropy trends for different prompts and resulting in frequent policy entropy fluctuations during training (see Fig.~\ref{fig_ab_geneval_curves}).
This shift reduces the diversity and visual quality of generated images, eventually leading to model degeneration. 
Similar observations regarding entropy collapse have also been reported in GRPO for text generation~\citep{cui2025entropyrl,wang2025_2080rule}.

\noindent\textbf{\textit{Solution}: Entropy reward for stabilized training.}
As discussed above, the reward model often produces hard signals without smooth variations among samples, causing confusion and unstable policy entropy. To mitigate this effect, we introduce a reward item based on policy entropy for the samples with the highest reward. Specifically, given the predicted token of sample $o_i$ distributions from current policy $\pi_\theta$ and reference policy $\pi_{\text{ref}}$, we compute their per-token entropy according to Eq.~\ref{eq_cal_entropy}, yielding $\mathcal{H}_\theta(o_i)$ and $\mathcal{H}_{\text{ref}}(o_i)$. The entropy reward is then defined as:
\begin{equation}
R^{\mathrm{ent}}_i \;=\; \bigl(1 + \bigl(\Delta\mathcal{H}_i\bigr)^2\bigr)^{-1}, \quad
\text{where}\ \Delta\mathcal{H}_i = \mathcal{H}_{\mathrm{ref}}(o_i) - \mathcal{H}_\theta(o_i);
\label{eq_cal_entropy_reward}
\end{equation}
\input{figs/vis_geneval_t2icompbench}
and is added to the original reward. Note that, to avoid potential influence on the final results, we apply the entropy reward only to the top-rewarded samples:
\begin{equation}
R_i^{\prime} = R_i + \lambda \cdot R^{\text{ent}}_i \cdot \mathbf{1}[R_i = \max_j R_j],
\label{eq_cal_total_reward}
\end{equation}
where $\lambda$ is a weighting coefficient and $\mathbf{1}[\cdot]$ is the indicator function. We set $\lambda$=0.4 in our experiments.
This entropy reward encourages the policy to maintain a level of uncertainty comparable to the reference model and prevents entropy collapse. As a result, it mitigates potential instability and reduces distributional drift relative to the reference policy (see Fig.~\ref{fig_ab_geneval_curves}). Training with the entropy reward yields markedly more stable entropy trajectories and substantially lower KL loss. Additional analysis of entropy reward can be found in Appendix~\ref{supp_add_analysis_entropy_reward}.

%% file: figs/motivation.tex
\begin{figure*}[t]
\setlength{\abovecaptionskip}{0.1cm}
\setlength{\belowcaptionskip}{0.1cm}
\begin{center}
\includegraphics[width=1\textwidth]{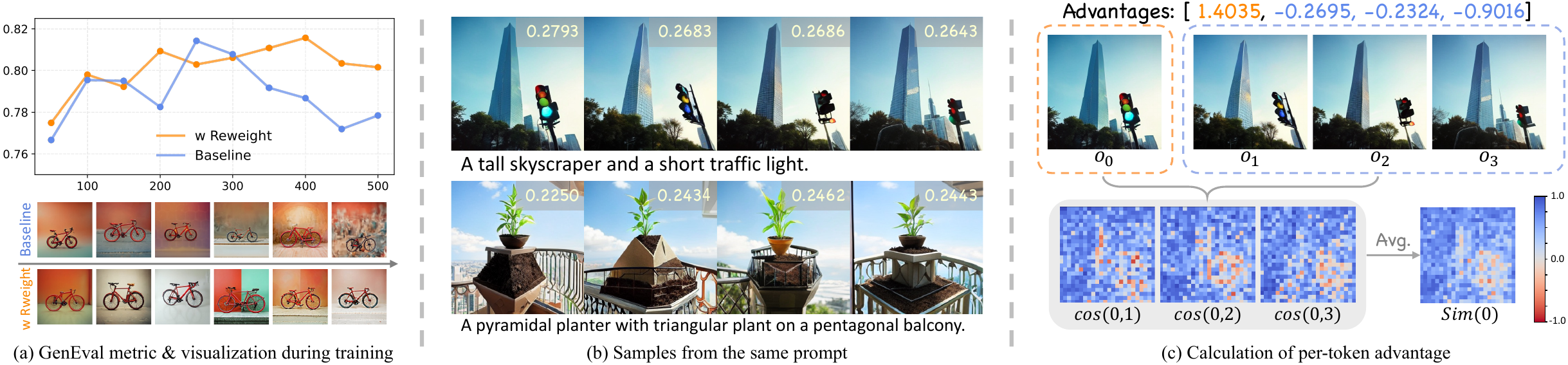}
\end{center}
\caption{
(a) Distributional disruption caused by conflicting gradients during training, especially with large learning rates and weak KL loss (lr = 5e-6, GenEval reward, KL weight = 0.01).
(b) Multiple samples generated from the same prompt exhibit high similarity, illustrated with Janus-pro 7B.
(c) Pairwise cosine similarity of VQ embeddings across images generated with the same prompt.
}
\vspace{-0.3cm}
\label{fig_motivation}
\end{figure*}

%% file: figs/motivation_entropy.tex


\begin{figure*}[t]
\setlength{\abovecaptionskip}{0.3cm}
\setlength{\belowcaptionskip}{0.3cm}
\centering
\begin{minipage}[t]{0.62\textwidth}
  \centering
  \includegraphics[width=\linewidth]{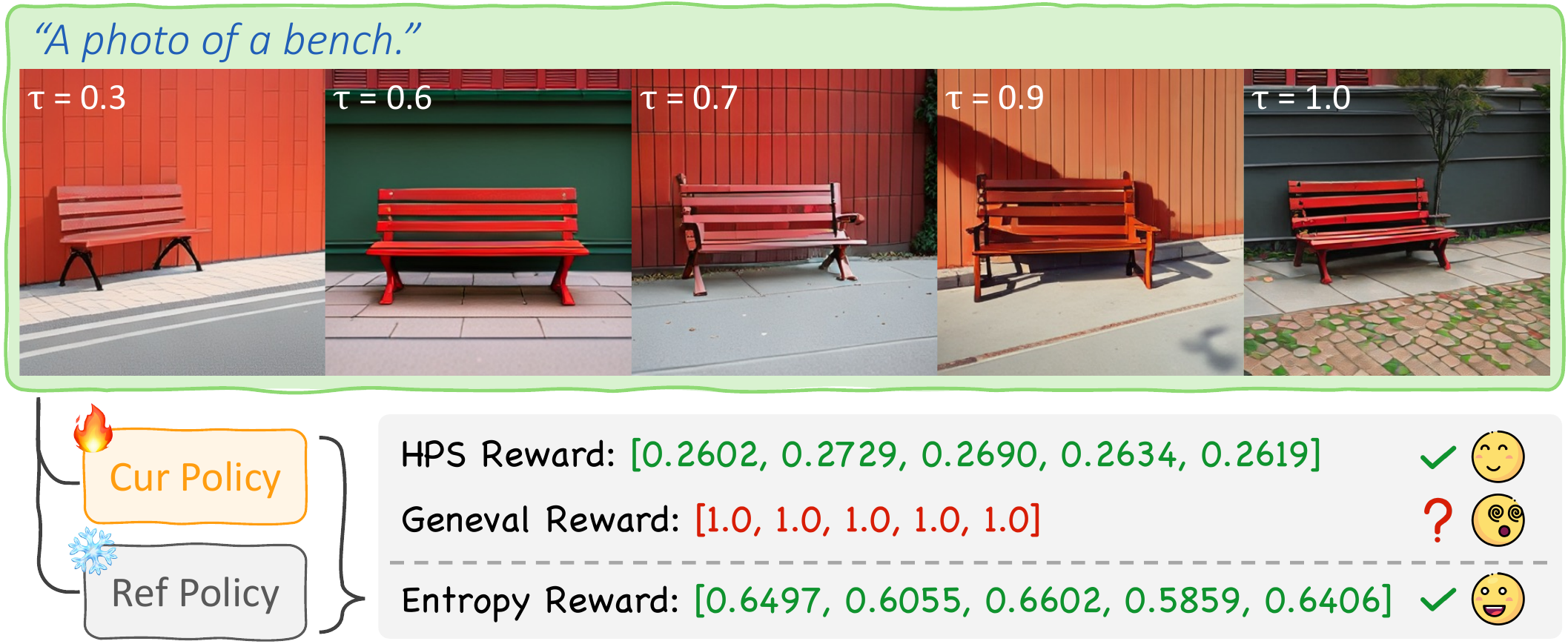}
  \captionof{figure}{
Samples from same prompt group generated by varying temperature $\tau$. HPS reward gives discriminative scores for high-quality images; rule-based GenEval reward often returns identical scores, causing confusion and instability; our entropy reward favors samples with policy entropy closer to the reference model, giving a clear preference.
  }
  \label{fig_motivation_entropy}
\end{minipage}%
\hfill
\begin{minipage}[t]{0.35\textwidth}
\setlength{\abovecaptionskip}{0.1cm}
\setlength{\belowcaptionskip}{0.1cm}
  \centering
    \vspace{-3.75cm}
  \includegraphics[width=\linewidth]{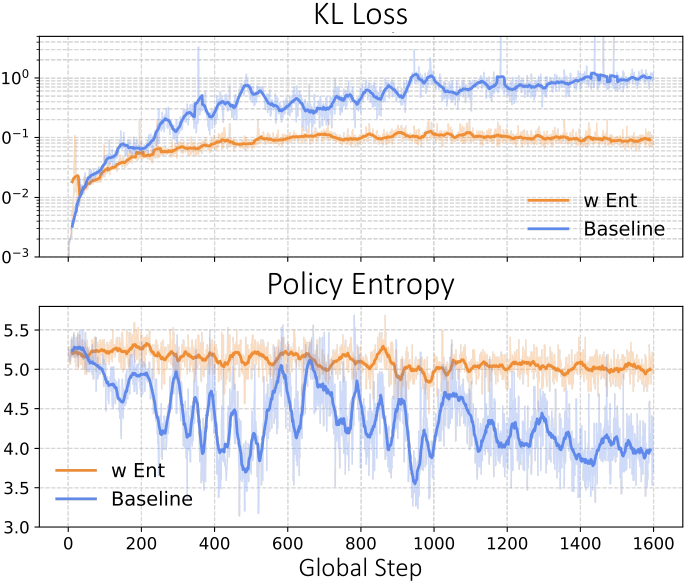}
  \captionof{figure}{
  During RL with GenEval reward, frequent fluctuations in policy entropy cause rapid KL loss growth $\&$ unstable training ("Baseline"). Entropy reward effectively alleviates this issue ("w Ent").
  }
  \label{fig_ab_geneval_curves}
\end{minipage}
\vspace{-3mm}
\end{figure*}

%% file: figs/vis_geneval_t2icompbench.tex
\begin{figure*}[t]
\setlength{\abovecaptionskip}{0.1cm}
\setlength{\belowcaptionskip}{0.1cm}
\begin{center}
\includegraphics[width=1\textwidth]{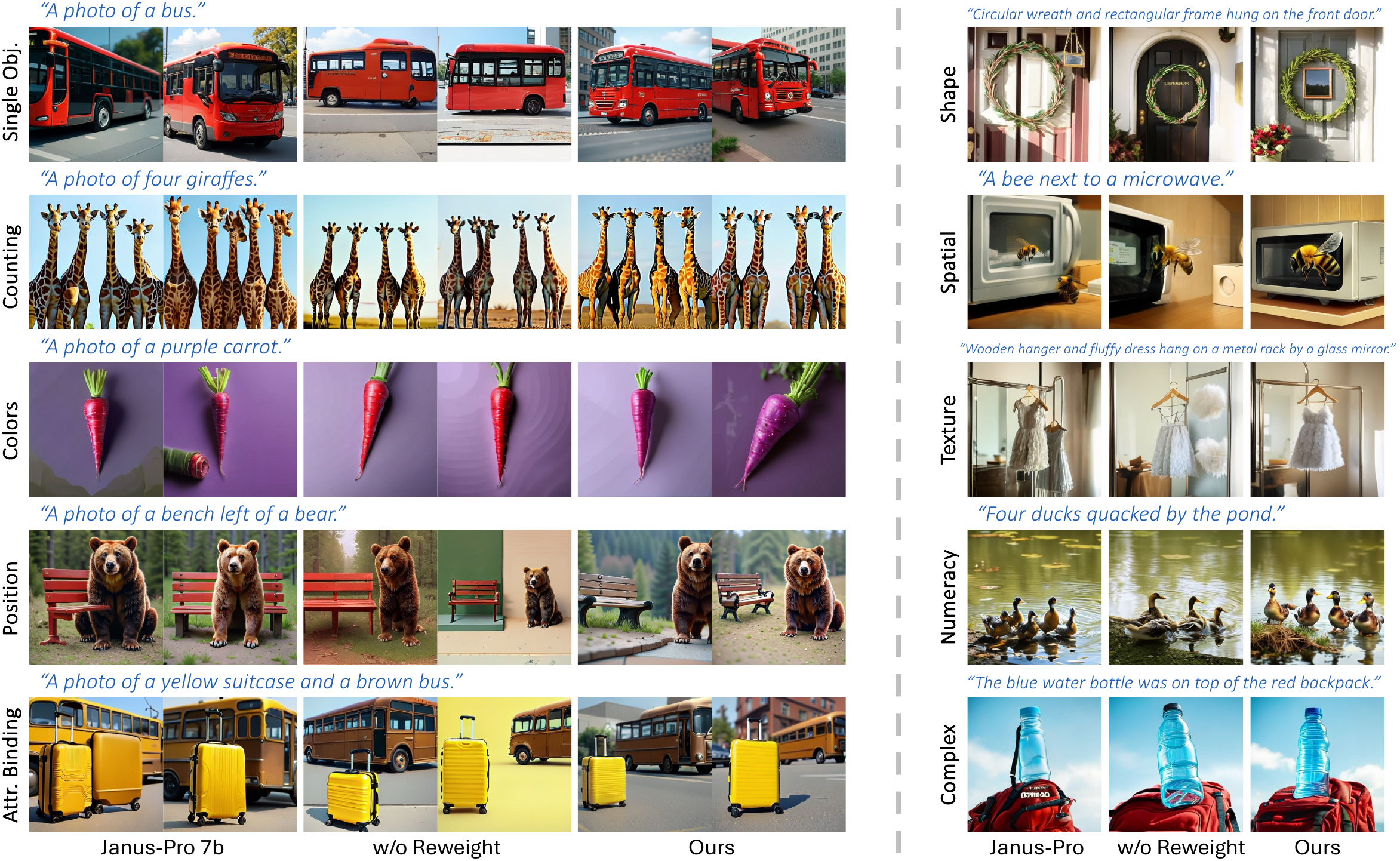}
\end{center}
\caption{
Generations comparison on GenEval (left) and T2I-CompBench (right) prompts. Compared to Baseline GRPO (with entropy reward to stablize training, labeled as ``w/o Reweight"), our advantage/KL reweighting preserves the base distribution and improves concept accuracy, yielding stronger structural stability, layout consistency, and finer detail than GRPO baselines.
}
\vspace{-0.3cm}
\label{fig_vis_geneval_t2icompbench}
\end{figure*}

%% file: sec/experiments.tex
\section{Experiment}
\label{experiment}
\subsection{Implementation Details}
\label{sec_implement_detail}
To systematically evaluate our method, we use Janus-Pro 7B~\citep{chen2025janus_pro} as our base model and GenEval rules as the reward. Images are generated at a resolution of 384 × 384 and CFG-scale is set to 5.
We set group size and batch size to 8, with initial learning rate of $5\times10^{-6}$ and KL coefficient $\beta=0.03$ to further stabilize training. 
We construct prompts proportionally to the base model's performance following~\citep{liu2025flow_grpo}, and train for 1,600 steps. 

Additionally, to verify the generality of our approach, we also conduct training with two additional reward types: (1) a mixture of human-preference, object-detector, and VQA rewards (HPS + Gdino~\citep{liu2024groundingdino} + Git~\citep{wang2022git}), following the T2I-R1 setup~\citep{jiang2025t2i_r1} to evaluate human preference and aesthetics; and (2) an OCR-based reward~\citep{gong2025seedream}, computed as the minimum edit distance between generated text and target text, to evaluate text-rendering capability. 
Full experimental details and parameter settings are provided in the Appendix~\ref{supp_add_imple_detail}.

\subsection{Main Results among Metrics}
We evaluate ours against baselines on metrics (GenEval, T2I-Compbench, etc.).
Note that ``Baseline" refers to vanilla GRPO, while ``Ours" denotes the method with proposed dynamic advantage/KL reweighting and entropy reward (the latter applied only for GenEval reward).

\noindent\textbf{GenEval.}
For GenEval reward, we provide both the last and best result and overall performance curves with respect to iteration (Table~\ref{tab_t2i_geneval} and Fig.~\ref{fig_vis_geneval_t2icompbench}).
The baseline GRPO improves the GenEval score from 0.78 to 0.86, while our method further boosts it to 0.89, surpassing many existing diffusion and AR models. In comparison, the T2I-R1 scheme brings only about 0.01 improvement on Janus-Pro.
\input{tables/table_geneval}

\noindent\textbf{T2I-Compbench.}
We report the generalization of models trained with the GenEval reward on T2I-Compbench; entries marked with ``${\dagger}$" in Table~\ref{tab_t2i_compbench} demonstrate that our method improves transfer performance. Additionally, the model trained with HPS+Gdino+Git mixed reward (``Baseline" and ``Ours") further demonstrates the stability improvement of structures and layouts in generated images. Additional visual comparison can be found in Appendix Fig.~\ref{fig_supp_add_visual_comp_t2icompbench}.

\noindent\textbf{HPS $\&$ ImageReward.}
For the HPS+Gdino+Git mixed reward, we report HPS and ImageReward metrics. Compared to the baseline without dynamic weighting, our method ("Ours") achieves higher scores and improved generation quality (see Table~\ref{table_hps_ocr} and Fig.~\ref{fig_vis_hps}). While vanilla GRPO enhances image-level details during training, dynamic weighting further refines the accuracy of rendered details and structure. Additional metrics (aesthetic, pickscore, DeQA) are given in Appendix Table~\ref{table_supp_imgrwd_t2i_r1}.

\input{figs/ours_geneval}
\input{tables/table_t2i_compbench}
\input{figs/vis_ocr}

\noindent\textbf{OCR.}
For OCR reward, we follow Flow-GRPO and evaluate on its OCR test set ($\sim$1,000 text-generation prompts) using average text edit distance (see Fig.~\ref{fig_vis_ocr} and Table~\ref{table_hps_ocr}). Ours yields more stable text generation, whereas baseline GRPO sometimes underperforms the original Janus-Pro in visual examples. The RL-trained model surpasses discrete AR and many diffusion models. 

\noindent\textbf{Generalization evaluation.}
We attribute the improved generalization to a better combination between base model's original distribution and RL update, which in turn avoids distribution shift and improves image quality. For models trained with GenEval reward, we evaluate effects of different training settings on T2I-Compbench, ImageReward and provide visualizations on GenEval (see Table~\ref{tab_t2i_compbench}, Table~\ref{table_generalize_geneval} and Fig.~\ref{fig_ab_geneval_vis}). Additional generalization experiments (e.g., model trained with GenEval and reward in T2I-R1~\citep{jiang2025t2i_r1} evaluated under human-preference metrics) are reported in Appendix~\ref{supp_add_experiment}. Removing the entropy reward or dynamic reweighting negatively affects image quality and generalization; by contrast, our method better balances performance and generality.

\subsection{Ablations \& Discussions}
\input{figs/ab_geneval}
\input{tables/table_ablation}
\noindent\textbf{Impact of advantage $\&$ KL reweighting.}
Fig.~\ref{fig_ours_geneval} shows the GenEval reward curve during training: compared to method without reweighting (``w/o Reweight") ours (``Ours") converges faster and consistently maintains above 0.88 in later steps. 
The dynamic weighting strategy provides clearer optimization directions, prevents distribution collapse, and yields images with finer details and better generalization, see Table~\ref{table_generalize_geneval} and Fig.~\ref{fig_vis_geneval_t2icompbench}, whereas baseline may suffer from distribution disruption during training, leading to degraded image quality and reduced generalization.

\noindent\textbf{Impact of entropy reward.}
As shown in Fig.~\ref{fig_ab_geneval_curves}, the entropy reward stabilizes the variations of KL loss and policy entropy during training, which protects the model distribution and further leads to performance improvements. Fig.~\ref{fig_ab_geneval} and Fig.~\ref{fig_ab_geneval_vis} presents the impact of including or removing the entropy reward (“w/o Ent”) on GenEval. It can be observed that removing the entropy reward results in noticeably reduced stability and larger fluctuations, thereby limiting further improvements in later-stage performance. More discussion about entropy reward can be found in Appendix~\ref{supp_add_analysis_entropy_reward}.

\noindent\textbf{Discussion of KL loss.}
During generation, KL loss is crucial for preserving original distribution of AR models. Removing KL loss may disrupt the distribution at early stages, preventing performance improvement.
DAPO~\citep{yu2025dapo} propose to discard samples with all-positive or all-negative rewards. Following this idea, we discard corresponding KL loss when reward standard deviation within a group equals zero. We find while this helps improve performance, it makes training more unstable (see ``w/o KL" in Fig.~\ref{fig_ab_geneval}) and may lead to degraded generation quality (see Fig.~\ref{fig_ab_geneval_vis}).


\noindent\textbf{Diversity of generated content.}
The proposed sample-similarity weighting may raise concerns regarding a potential reduction in diversity of generated images. However, results on GenEval reward indicate that, by better preserving original distribution, the weighting can in fact enhance diversity in certain cases. Nevertheless, instances of decreased diversity do exist (see Fig.~\ref{fig_vis_diversity}). A more detailed discussion on diversity is provided in the Appendix~\ref{supp_add_analysis_diversity}.

\input{figs/ab_geneval_curves}

%% file: tables/table_geneval.tex
\begin{table}[]
\centering
\setlength{\abovecaptionskip}{0.1cm}
\setlength{\belowcaptionskip}{0.1cm}
\caption{
Quantitative results on GenEval across models. Compared to vanilla GRPO (``Baseline"), both the entropy reward (``+Ent") and advantage/KL reweighting (``Ours") improve performance; we provide GenEval value of our last checkpoint (``Ours") and best (``Ours*") for reference.
}
\setlength{\tabcolsep}{3.5mm}{
\resizebox{1\columnwidth}{!}{%
\begin{tabular}{lccccccc}
\toprule
\multicolumn{1}{l}{Model} & Overall$\uparrow$ & Single Obj.$\uparrow$ & Two Obj.$\uparrow$ & Counting$\uparrow$ & Colors$\uparrow$ & Position$\uparrow$ & Attr. Binding$\uparrow$ \\ \midrule
Pixart-$\alpha$~\citeyearpar{chen2023pixartalphafasttrainingdiffusion} &0.48 & 0.98	& 0.50	& 0.44 & 0.80 & 0.08 & 0.07 \\
SD3~\citeyearpar{esser2024scaling}& 0.74 & \best{0.99} & 0.94 & 0.72 & 0.89 & 0.33 & 0.60 \\
FLUX.1-dev~\citeyearpar{flux2024}& 0.66 & 0.98 & 0.79 & 0.73 & 0.77 & 0.22 & 0.45 \\
Sana-1.5~\citeyearpar{xie2025sana}& 0.81 & \best{0.99} & 0.93 & \best{0.86} & 0.84 & 0.59 & 0.65 \\ \midrule
LlamaGen~\citeyearpar{sun2024llamagen}& 0.32 & 0.71 & 0.34 & 0.21 & 0.58 & 0.07 & 0.04 \\
Show-o~\citeyearpar{xie2024show_o}& 0.68 & 0.98 & 0.80 & 0.66 & 0.84 & 0.31 & 0.50 \\
Infinity~\citeyearpar{han2024infinity}& 0.73 &   -  & 0.85 &   -  &   -  & 0.49 & 0.57 \\
GPT-4o~\citeyearpar{gpt4o}& 0.85 & \best{0.99} & 0.92 & 0.85 & \best{0.91} & 0.75 & 0.66 \\ \midrule
Janus-Pro-7B~\citeyearpar{chen2025janus_pro}& 0.78    & 0.98        &  0.86    & 0.56     & 0.89   & 0.76     & 0.63          \\
T2I-R1~\citeyearpar{jiang2025t2i_r1}                    & 0.79    & \best{0.99}        &  0.91    & 0.53     & \best{0.91}   & 0.76     & 0.65          \\
Baseline                  & 0.86  & 0.97 & 0.92 & 0.82 & 0.86 & 0.84 & 0.72          \\
+Ent                  & 0.87    & \best{0.99}        &  0.94    & 0.78     & 0.90   & 0.89     & 0.73          \\
Ours                      & 0.88    & \best{0.99}        & 0.93     & 0.82     & 0.89   & \best{0.91}     &  0.77          \\ 
Ours*                      & \best{0.89}    & \best{0.99}        & \best{0.95}     & 0.82     & 0.90   & 0.89     & \best{0.79}          \\ \bottomrule
\end{tabular}

}}
\label{tab_t2i_geneval}
\vspace{-3mm}
\end{table}

%% file: figs/ours_geneval.tex
\begin{figure*}[t]
\setlength{\abovecaptionskip}{0.1cm}
\setlength{\belowcaptionskip}{0.1cm}
\centering
\begin{minipage}[t]{0.37\textwidth}
  \centering
  \includegraphics[width=\linewidth]{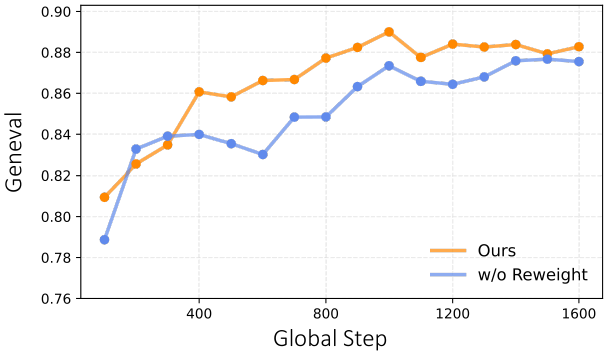}
  \captionof{figure}{GenEval vs. global steps during RL. Ours converges faster and attains higher final performance than baseline GRPO with entropy reward.}
  \label{fig_ours_geneval}
\end{minipage}%
\hfill
\begin{minipage}[t]{0.6\textwidth}
  \centering
  \includegraphics[width=\linewidth]{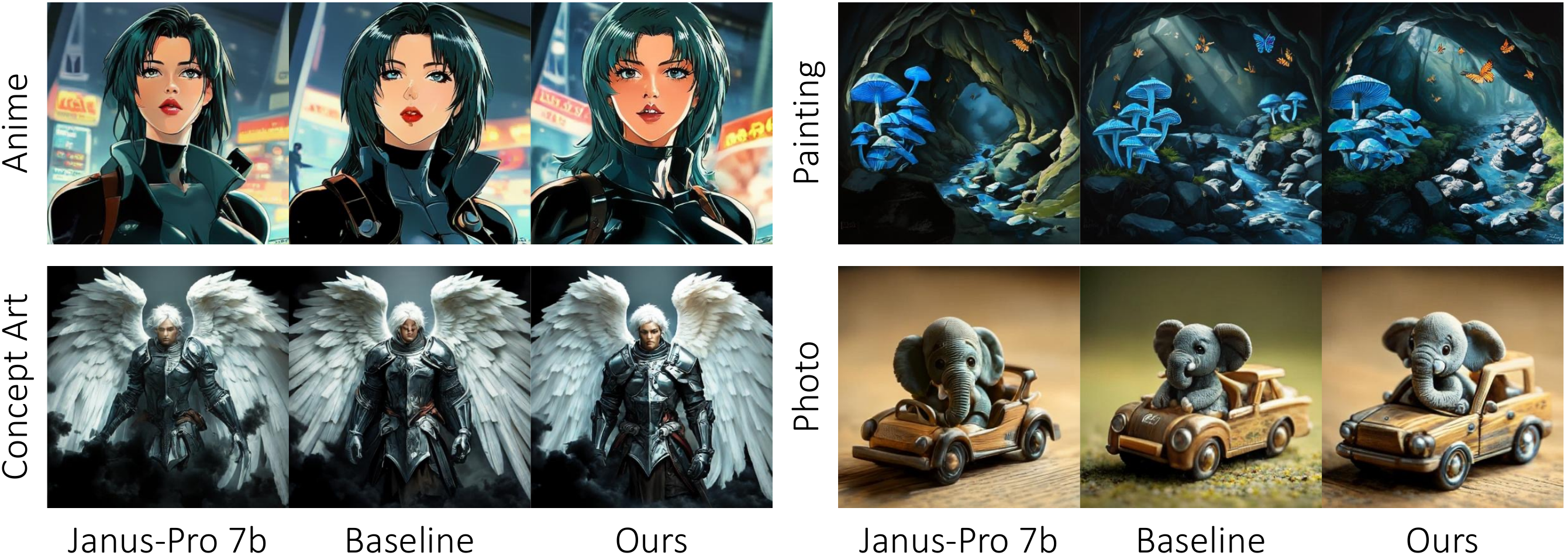}
  \captionof{figure}{Visualize comparison on HPS prompts. Our method delivers clearer fine details and stronger structural stability than both the baseline GRPO and the original Janus-Pro 7b (see human face, armors and structure of cars).}
  \label{fig_vis_hps}
\end{minipage}
\vspace{-3mm}
\end{figure*}

%% file: tables/table_t2i_compbench.tex
\begin{table}[t]
\vspace{-1mm}
\centering
\setlength{\abovecaptionskip}{0.1cm}
\setlength{\belowcaptionskip}{0.1cm}
\caption{
Quantitative results on T2I-compbench. 
Compared to vanilla GRPO (``Baseline"), our models trained with the GenEval reward generalize better, yielding consistent gains across multiple sub-benchmarks. ``Spat." is short for spatial. ``${\dagger}$" means model is trained on GenEval reward.
}
\setlength{\tabcolsep}{2.5mm}{
\resizebox{1.0\columnwidth}{!}{%
\begin{tabular}{lcccccccc}
\specialrule{1.2pt}{0pt}{0pt} 
\multicolumn{1}{l}{Model} & Color$\uparrow$ & Shape$\uparrow$ & Texture$\uparrow$ & 2D-Spat.$\uparrow$ & 3D-Spat.$\uparrow$ & Non-Spat.$\uparrow$ & Numeracy$\uparrow$ & Complex$\uparrow$ \\ 
\midrule
SD3~\citeyearpar{esser2024scaling}& \second{0.8094} & \second{0.5864} & 0.7297 & 0.3219 &  \second{0.4044}   & \best{0.3143}  & 0.6078  & 0.3780 \\
FLUX.1-dev~\citeyearpar{flux2024}& 0.7407 & 0.5718 & 0.6922 & 0.2863 & 0.3866 & 0.3127 & \second{0.6185} & -        \\ 
Sana-1.5~\citeyearpar{xie2025sana}& 0.7625 & 0.5426 & 0.6761 & \best{0.3814} & \best{0.4088} & 0.3123  & 0.6110 & 0.3727 \\ 
\midrule
LlamaGen~\citeyearpar{sun2024llamagen}& 0.2996 & 0.3212 & 0.3888 &  0.1004 & 0.1530  & 0.2729  & 0.2747 & 0.2501   \\
Show-o~\citeyearpar{xie2024show_o}& 0.7327 & 0.5264 & 0.6815 & 0.3697 & 0.3996 & 0.3106 & \best{0.6209} & 0.3572   \\ 
Infinity~\citeyearpar{han2024infinity}& 0.7379 & 0.4650 & 0.5919  & 0.2215 & 0.3846 & 0.3076 & 0.5475 & 0.3689  \\
\midrule
Janus-Pro-7B~\citeyearpar{chen2025janus_pro}& 0.6355   & 0.3494  & 0.4929  & 0.1931   & 0.3279 & 0.3087    & 0.4423 & 0.3566  \\
T2I-R1~\citeyearpar{jiang2025t2i_r1}& \best{0.8130}   & 0.5852  & 0.7243  & 0.3378   & -        & 0.3090    & -       & 0.3993  \\
Baseline$^{\dagger}$   & 0.7143  & 0.4028 & 0.6085 & 0.2763  & 0.3692  & 0.3090   & 0.4394 & 0.3586 \\
Ours$^{\dagger}$       & 0.7463 & 0.4388 & 0.6443 & 0.3053  & 0.3667  & 0.3107   & 0.5278 & 0.3779 \\ 
Baseline               & 0.7829  & 0.5842 & \second{0.7380} & 0.3674  & 0.4042  & 0.3131   & 0.5902 & \best{0.4004} \\
Ours                   & 0.7842 & \best{0.5923} & \best{0.7451} & \second{0.3731}  & 0.4005  & \second{0.3136}   & 0.5993 & \second{0.3997} \\ 
\specialrule{1.2pt}{0pt}{0pt} 
\end{tabular}
}}
\label{tab_t2i_compbench}
\vspace{-4mm}
\end{table}

%% file: figs/vis_ocr.tex
\begin{figure*}[b]
\vspace{-4mm}
\centering
\begin{minipage}[t]{0.6\textwidth}
  \centering
  \includegraphics[width=\linewidth]{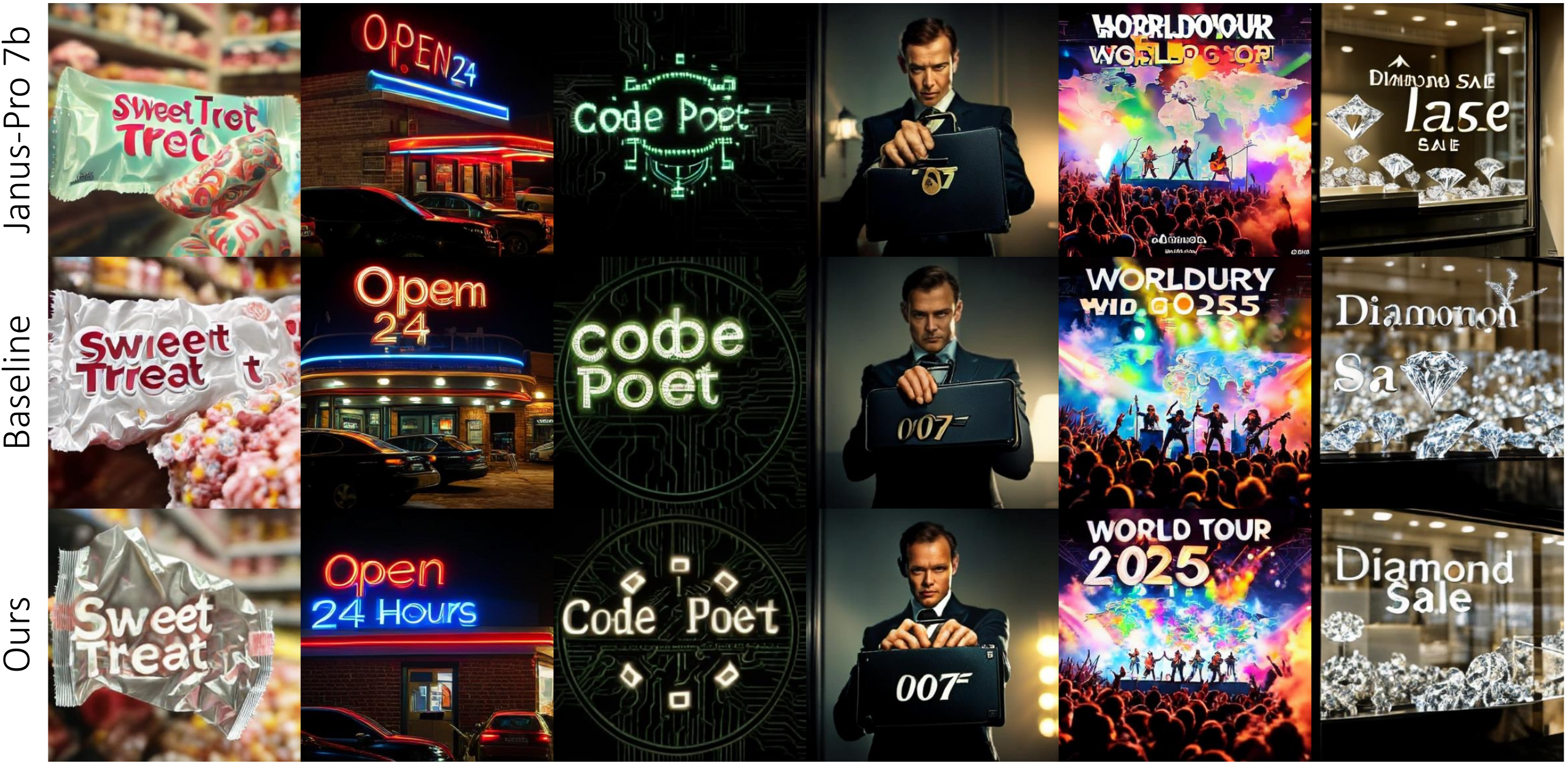}
  \captionof{figure}{
Visualization of text rendering capability. Compared with the baseline, ours more accurately captures the textual structure while maintaining image generation quality.
  }
  \label{fig_vis_ocr}
\end{minipage}%
\hfill
\begin{minipage}[t]{0.37\textwidth}
\vspace{-4cm}
\input{tables/table_hps_ocr}
\end{minipage}
\end{figure*}

%% file: tables/table_hps_ocr.tex
\centering
\setlength{\abovecaptionskip}{0.1cm}
\setlength{\belowcaptionskip}{0.1cm}
\captionof{table}{
Quantitative evaluation of baseline GRPO on and ours on HPS, ImageReward, and text rendering.
}
\setlength{\tabcolsep}{1.2mm}{
\resizebox{1\columnwidth}{!}{%
\begin{tabular}{@{}lccc@{}}
\toprule
\multicolumn{1}{c}{Model}                                 & HPS$\uparrow$ & ImgRwd$\uparrow$ & OCR$\uparrow$ \\ \midrule
Pixart-$\alpha$~\citeyearpar{chen2023pixartalphafasttrainingdiffusion} & 30.76 & 0.75 & 0.04 \\
SD3~\citeyearpar{esser2024scaling}     & 30.22 & 1.00 & 0.57    \\
FLUX.1-dev~\citeyearpar{flux2024}   & 31.35 & 1.10 & 0.63    \\
Sana-1.5~\citeyearpar{xie2025sana}     & 30.36 & 1.08 & 0.33    \\ \midrule
LlamaGen~\citeyearpar{sun2024llamagen} & 23.92 & -0.36 & 0.04    \\
Show-o~\citeyearpar{xie2024show_o}    & 27.98 & 0.86 & 0.08    \\
Infinity~\citeyearpar{han2024infinity} & 30.60 & 0.88 & 0.36    \\ \midrule
Janus-Pro-7B~\citeyearpar{chen2025janus_pro} & 28.64 & 0.76 & 0.21    \\
T2I-R1~\citeyearpar{jiang2025t2i_r1}       & 29.83 & 0.94  & 0.23    \\
Baseline     & 29.73 & 0.93 & 0.46    \\
Ours         & 29.87 & 0.98 & 0.49    \\ \bottomrule
\end{tabular}
}}
\label{table_hps_ocr}

%% file: figs/ab_geneval.tex
\begin{figure*}[t]
\setlength{\abovecaptionskip}{0.1cm}
\setlength{\belowcaptionskip}{0.1cm}
\begin{center}
\includegraphics[width=1\textwidth]{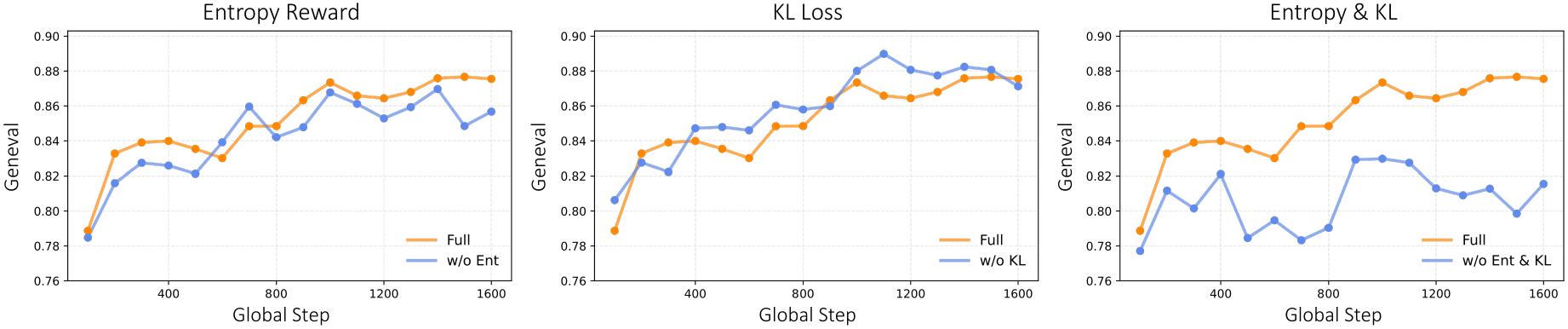}
\end{center}
\caption{
Ablations on Geneval: from a Baseline+Entropy reward setting (``Full"), dropping the KL on zero-variance groups (“w/o KL”) or the entropy reward (“w/o Ent”) reduces training stability.
}
\vspace{-0.3cm}
\label{fig_ab_geneval}
\end{figure*}

%% file: tables/table_ablation.tex
\begin{table}[t]
\vspace{-1mm}
\centering
\setlength{\abovecaptionskip}{0.1cm}
\setlength{\belowcaptionskip}{0.1cm}
\caption{
Ablations across variants on GenEval (in-distribution), T2I-CompBench and ImageReward (out-of-distribution). Ours achieves the best in- and out-of-distribution performance over baselines.
}
\setlength{\tabcolsep}{1.5mm}{
\resizebox{1.0\columnwidth}{!}{%
\begin{tabular}{@{}l|ccccccc|ccccc|c@{}}
\toprule
\multirow{2}{*}{Model} & \multicolumn{7}{c|}{GenEval}                           & \multicolumn{5}{c|}{T2I-Compbench}          & \multirow{2}{*}{ImgRwd} \\ \cmidrule(lr){2-13}
                       & Overall & Single & Two  & Count & Color & Pos. & Attr & Color  & Shape  & Text.  & 2d Spat. & Num.   &                         \\ \midrule
Janus-Pro 7B           & 0.78    & 0.98   & 0.86 & 0.56  & \second{0.89}  & 0.76 & 0.63  & 0.6355 & 0.3494 & 0.4929 & 0.1931  & 0.4423 & 0.76                    \\
Ours                   & \best{0.88}    & \best{0.99}   & \second{0.93} & \best{0.82}  & \second{0.89}  & \best{0.91} & \best{0.77}  & \best{0.7463} & \best{0.4388} & \second{0.6443} & \second{0.3053}  & \best{0.5278} & \best{0.80}                    \\
w/o Reweight (Full)    & \second{0.87}    & \best{0.99}   & \best{0.94} & 0.78  & \best{0.90}  & \second{0.89} & 0.73  & 0.7241 & 0.4063 & 0.6032 & 0.2788  & 0.4771 & 0.74                    \\
w/o Ent (Baseline)     & 0.86    & 0.97   & 0.92 & \best{0.82}  & 0.86  & 0.84 & 0.72  & 0.7143 & 0.4028 & 0.6085 & 0.2763  & 0.4394 & 0.67                    \\
w/o KL                 & \second{0.87}    & 0.98   & \second{0.93} & \best{0.82}  & 0.86  & 0.86 & \second{0.74}  & 0.7214 & 0.4149 & 0.6214 & 0.3005  & \second{0.4923} & \second{0.77}                    \\
w/o Ent \& KL          & 0.81    & 0.98   & 0.90 & 0.73  & 0.88  & 0.71 & 0.67  & \second{0.7269} & \second{0.4250} & \best{0.6558} & \best{0.3072}  & 0.4591 & 0.76                    \\ \bottomrule
\end{tabular}
}}
\label{table_generalize_geneval}
\vspace{-3mm}
\end{table}

%% file: figs/ab_geneval_curves.tex
\begin{figure*}[ht]
\setlength{\abovecaptionskip}{0.1cm}
\setlength{\belowcaptionskip}{0.1cm}
\centering
\begin{minipage}[t]{0.6\textwidth}
  \centering
  \includegraphics[width=\linewidth]{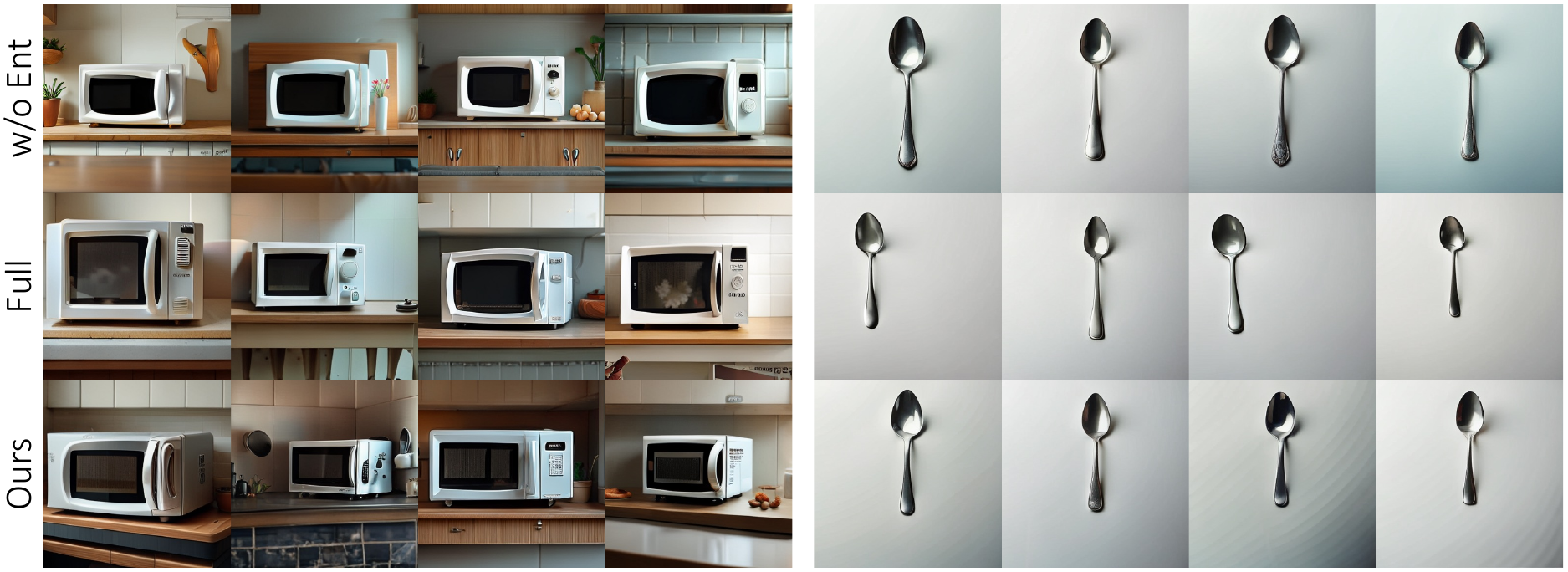}
  \captionof{figure}{
  Visualization of diversity. In some cases, the ``Full" (Baseline+Entropy reward) and ``w/o KL" reduce diversity due to distribution drift. However, ``Ours" may risk lowered diversity when foreground–background contrast is large.
  }
  \label{fig_vis_diversity}
\end{minipage}%
\hfill
\begin{minipage}[t]{0.37\textwidth}
  \centering
  \includegraphics[width=\linewidth]{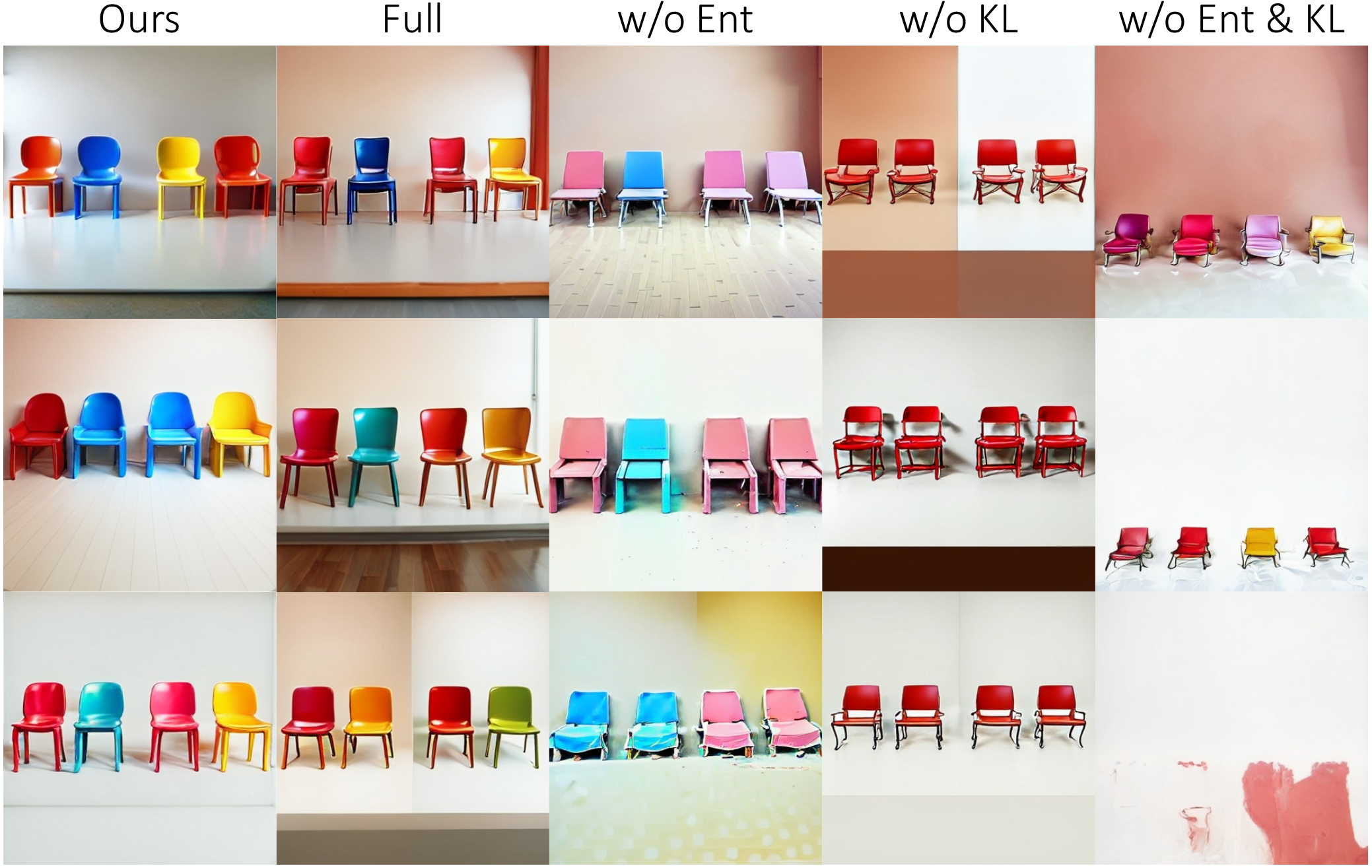}
  \captionof{figure}{
Removing entropy or zeroing KL for equal-reward groups worsens images compared to Ours and Baseline+Entropy reward (``Full").
  }
  \label{fig_ab_geneval_vis}
\end{minipage}
\vspace{-3mm}
\end{figure*}

%% file: sec/conclusion.tex
\section{Conclusion}
In this work, we investigate limitations of existing GRPO  for AR text-to-image method and propose dynamic weighting strategy based on characteristics of image tokens and AR image generation to avoid redundant gradients that perturb the model’s learned distribution. We further introduce an entropy-based reward to stabilize training. Together, these techniques improve training stability and generalizability, raising the performance ceiling of AR generative models. We hope this work inspires further research toward closing the gap between AR and diffusion-based methods.

%% file: sec/supp/add_method_description.tex
\section{Additional Method Description}

In one GRPO iteration, the current policy first generates a group of samples $\{o_1,o_2,\ldots\}$. The generated samples are (i) scored by the reward rule to obtain sequence-level rewards $\{R_1,R_2,\ldots\}$ and (ii) fed to both the current policy and the reference policy to obtain the corresponding log-probabilities and the two policy entropies $H_\theta$ and $H_{\text{ref}}$. We compute the entropy reward according to Eq.~\ref{eq_cal_entropy_reward} and the combined reward $R'$ according to Eq.~\ref{eq_cal_total_reward}. After normalization we obtain the advantages $\hat{A}$, which are reweighted into $\tilde{A}$ and KL coefficients following Eqs.~\ref{eq_embed_sim}$\sim$\ref{eq_embed_advantage_reweight}. Using the log-probabilities and the KL schedule we compute the importance ratios $r_{i,t}(\theta)$ as in Eq.~\ref{eq_important_ratio}, and finally form the GRPO objective $J_{\text{GRPO}}(\theta)$ for the current update using Eq.~\ref{eq_grpo_opt_objective}. A brief visualization of our framework can be found in Fig.~\ref{fig_supp_framework}.

\input{figs/supp/supp_framework}

%% file: figs/supp/supp_framework.tex
\begin{figure*}[h]
\setlength{\abovecaptionskip}{0.1cm}
\setlength{\belowcaptionskip}{0.1cm}
\begin{center}
\includegraphics[width=1\textwidth]{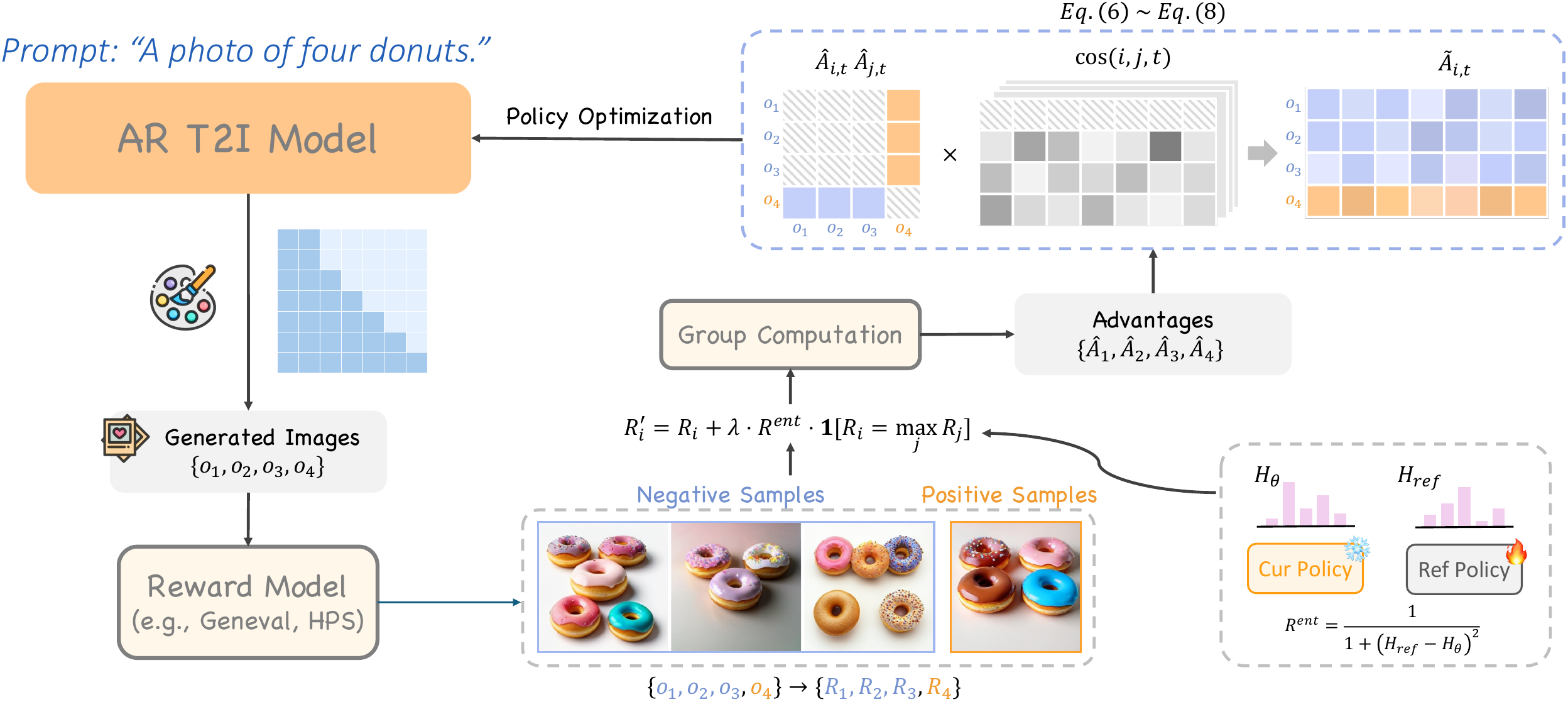}
\end{center}
\caption{
Overall view of our framework. At each iteration the policy generates a group of samples ${o_i}$ that are scored to yield sequence rewards ${R_i}$ and evaluated by the current and reference policies to obtain log-probs and entropies $H_\theta,H_{\rm ref}$. The entropy reward and sequence reward combine into $R'$, which is normalized to advantages $\hat A$, reweighted to $\tilde A$ with token KL weights, and used with importance ratios $r_{i,t}(\theta)$ to form the GRPO objective $J_{\rm GRPO}(\theta)$.
}
\vspace{-0.3cm}
\label{fig_supp_framework}
\end{figure*}

%% file: sec/supp/detailed_exp_setting.tex
\section{Detailed Experimental Setting}
\subsection{Implementation Details}
\label{supp_add_imple_detail}
\noindent\textbf{Training pipeline}
Following the setup of T2I-R1, we adopt Janus-Pro 7B as the base model. Images are generated at a resolution of $384\times384$. The batch size is set to 8 (i.e., 8 prompts per step), with a group size of 8 (8 images per prompt). During GRPO inference, we apply classifier-free guidance (CFG) with a scale of 5, consistent with the official Janus-Pro configuration, and the sampling temperature is fixed to 1. The training is conducted on 8 H100 GPUs using DeepSpeed ZeRO-3 and the HuggingFace Transformers library, and Adam optimizer is used with $\beta_1=0.9$ and $\beta_2=0.999$.

\noindent\textbf{Reward function}
We evaluate several types of reward functions:

1) Geneval reward: Following Flow-GRPO~\citep{liu2025flow_grpo}, we adopt Geneval rules~\citep{ghosh2023geneval} for scoring. Rewards are defined according to task type: for counting, $r = 1 - |N_{\text{gen}} - N_{\text{ref}}| / N_{\text{ref}}$; for color and position constraints, rewards reflect the proportion of correctly matched objects, with full match giving $1$ and partial mismatch proportionally reduced. The final reward is averaged over all clauses.
Geneval reward requires a larger learning rate and KL penalty to achieve stable improvements. Therefore, we set $\text{lr}=5\times10^{-6}$, $\beta=0.03$, train for 1600 steps, and disable gradient accumulation to enable faster and more stable performance gains.

2) Combination reward: T2I-R1~\citep{jiang2025t2i_r1} adopts a combination of rewards, including human preference score (HPS), object detector (GroundingDINO), and visual question answering model (GIT), linearly combined together. We follow their setting with learning rate $1\mathrm{e}{-6}$, $\beta=0.01$, gradient accumulation=2, and total training steps=1600, to achieve more stable performance improvement.

3) OCR reward: Following Flow-GRPO~\citep{liu2025flow_grpo}, given a prompt, we generate images and apply an existing OCR tool—specifically, PaddleOCR—to compute the minimum edit distance $N_e$ between the rendered text and the target text. The corresponding reward is then defined as $r = \max(1 - N_e / N_{\text{ref}}, 0)$, where $N_{\text{ref}}$ is the number of characters inside the quotation marks in the prompt. We set $\text{lr}=1\times10^{-6}$, $\beta=0.01$, training for 1,600 steps and no gradient accumulation for OCR reward.

\noindent\textbf{Data construction}
1) For the Geneval benchmark, following Flow-GRPO, we adopt the Geneval-style evaluation prompts. Training data are constructed according to Janus-Pro’s accuracy on different categories, with the ratio single\_object:two\_object:counting:colors:position:color\_attr set to 0:1:7:1:4:5. 
2) For the mixed reward setting, we follow T2I-R1 and use the same prompts, consisting of 6k+ text prompts with GPT-4o-mini extracted objects and attributes from T2I-CompBench~\citep{huang2023t2i_compbench} and~\cite{guo2025imagecot}.
3) For the OCR reward, we use the training and test sets provided by Flow-GRPO, which consist of raw image prompts containing text renderings generated by GPT-4o, includes 20K training prompts and 1K test prompts.

%% file: sec/supp/add_metric.tex
\section{Additional Quantitative Results}
\subsection{Additional Experiments}
\label{supp_add_experiment}
\noindent\textbf{Generalization experiments with GenEval reward}.
Here we provide additional generalization experiments on the GenEval reward. Specifically, we compute several image quality–related metrics on the HPS prompts and DrawBench prompts (see Table~\ref{table_supp_imgrwd_geneval}). The original GRPO algorithm tends to collapse during training, which harms the quality of generated images and even yields lower scores than the original Janus-Pro on certain metrics. Adding an entropy reward helps stabilize training and further improves image quality. In addition, our proposed similarity-based dynamic weighting scheme also contributes to image quality improvements.

\input{tables/supp/table_imgrwd_geneval}

\begin{figure*}[t]
\setlength{\abovecaptionskip}{0.1cm}
\setlength{\belowcaptionskip}{0.1cm}
\begin{center}
\includegraphics[width=0.7\textwidth]{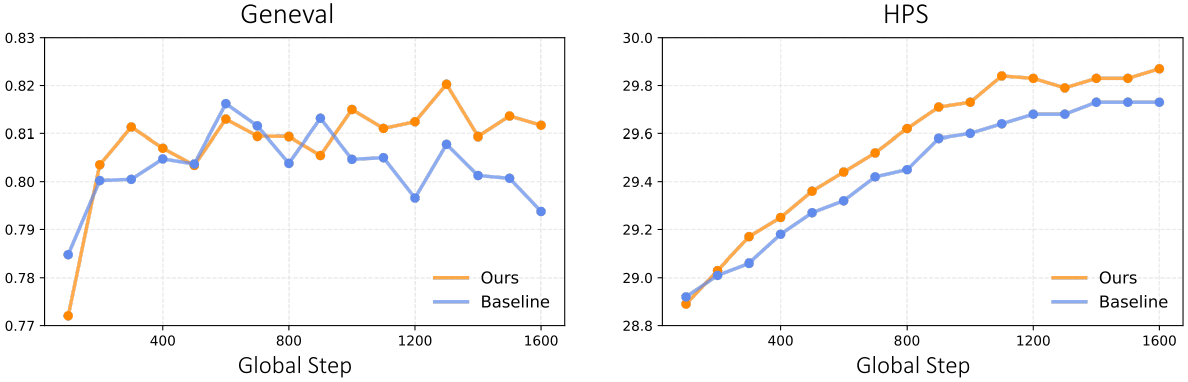}
\end{center}
\caption{
Evolution of HPS and GenEval metrics during training with HPS+Gdino+Git rewards. Compared to the baseline, our method achieves faster HPS improvement, more stable GenEval gains, and no noticeable late-stage decline.
}
\vspace{-0.3cm}
\label{fig_supp_ab_hps_git_gdino}
\end{figure*}

\noindent\textbf{Metrics during training with reward on~\citep{jiang2025t2i_r1}}.
Fig.~\ref{fig_ab_geneval} in the main text shows the evolution of GenEval metrics during training with the GenEval reward. Here, we provide the GenEval and HPS metrics over global steps using the hyperparameters from ~\citep{jiang2025t2i_r1} (Fig.~\ref{fig_supp_ab_hps_git_gdino}). Compared to the baseline, dynamic weighting accelerates the improvement of HPS and stabilizes GenEval metrics, preventing the late-stage decline observed in the baseline.

\noindent\textbf{Generalization experiments with reward on~\citep{jiang2025t2i_r1}}. In the main text, we reported results on T2I-CompBench and a subset of image-quality metrics (HPS, ImageReward, etc.). Here we additionally provide the corresponding GenEval scores to further substantiate the generalization of the proposed method; see Table~\ref{table_supp_geneval_generalize} and Table~\ref{table_supp_imgrwd_t2i_r1}.
Compared with the baseline model, the dynamic similarity weighting strategy helps improve image quality.

\input{tables/supp/table_geneval_generalize}
\input{tables/supp/table_imgrwd_t2i_r1}

\input{sec/supp/adapt_add_models}

%% file: tables/supp/table_imgrwd_geneval.tex
\begin{table}[h]
\centering
\setlength{\abovecaptionskip}{0.1cm}
\setlength{\belowcaptionskip}{0.1cm}
\caption{
Generalization on Geneval benchmark for models trained with the Geneval reward on HPS and drawbench.
}
\setlength{\tabcolsep}{3mm}{
\resizebox{0.7\columnwidth}{!}{%
\begin{tabular}{@{}lccccc@{}}
\toprule
\multicolumn{1}{c}{} & HPS$\uparrow$   & ImageReward$\uparrow$ & Pickscore$\uparrow$ & DeQA$\uparrow$ & Aesthetic$\uparrow$ \\ \midrule
Janus-Pro-7b         & 28.64 & 0.76        & 21.83     & 3.53 & 5.68      \\
T2I-R1               & 29.83 & 0.94        & 22.03     & 3.65 & 5.91      \\
Baseline             & 28.49 & 0.67        & 21.86     & 3.49 & 5.55      \\
+Ent                 & 28.72 & 0.74        & 21.87     & 3.53 & 5.65      \\
Ours                 & 28.72 & 0.80        & 21.91     & 3.54 & 5.65      \\ \bottomrule
\end{tabular}
}}
\label{table_supp_imgrwd_geneval}
\vspace{-3mm}
\end{table}

%% file: tables/supp/table_geneval_generalize.tex
\begin{table}[h]
\centering
\setlength{\abovecaptionskip}{0.1cm}
\setlength{\belowcaptionskip}{0.1cm}
\caption{
Generalization on GenEval for models trained with the HPS + GIT + Grounding DINO reward on T2I CompBench style prompts. The baseline yields almost no gain on the GenEval score of Janus Pro, whereas our method under the same setting increases GenEval from 0.78 to 0.81.
}
\setlength{\tabcolsep}{3mm}{
\resizebox{1\columnwidth}{!}{%
\begin{tabular}{@{}lccccccc@{}}
\toprule
\multicolumn{1}{l}{Model} & Overall$\uparrow$ & Single Obj.$\uparrow$ & Two Obj.$\uparrow$ & Counting$\uparrow$ & Colors$\uparrow$ & Position$\uparrow$ & Attr. Binding$\uparrow$ \\ \midrule
Janus-Pro-7B~\citeyearpar{chen2025janus_pro}& 0.78    & 0.98        &  0.86    & 0.56     & 0.89   & 0.76     & 0.63          \\
T2I-R1~\citeyearpar{jiang2025t2i_r1}                    & 0.79    & 0.99        &  0.91    & 0.53     & 0.91   & 0.76     & 0.65          \\
Baseline                  & 0.79  & 0.98 & 0.86 & 0.57 & 0.86 & 0.83 & 0.64          \\
Ours                      & 0.81    & 0.98        & 0.90     & 0.62     & 0.88   & 0.83     & 0.64          \\ \bottomrule
\end{tabular}
}}
\label{table_supp_geneval_generalize}
\vspace{-3mm}
\end{table}

%% file: tables/supp/table_imgrwd_t2i_r1.tex
\begin{table}[h]
\centering
\setlength{\abovecaptionskip}{0.1cm}
\setlength{\belowcaptionskip}{0.1cm}
\caption{
Evaluation for models trained with the HPS + GIT + Grounding DINO reward on drawbench.
}
\setlength{\tabcolsep}{3mm}{
\resizebox{0.6\columnwidth}{!}{%
\begin{tabular}{@{}lcccc@{}}
\toprule
\multicolumn{1}{c}{} & ImageReward$\uparrow$ & Pickscore$\uparrow$ & DeQA$\uparrow$ & Aesthetic$\uparrow$ \\ \midrule
Janus-Pro-7b         & 0.76        & 21.83     & 3.53 & 5.68      \\
T2I-R1               & 0.94        & 22.03     & 3.65 & 5.91      \\
Baseline             & 0.93        & 22.07     & 3.65 & 5.83      \\
Ours                 & 0.98        & 22.09     & 3.69 & 5.89      \\ \bottomrule
\end{tabular}
}}
\label{table_supp_imgrwd_t2i_r1}
\vspace{-3mm}
\end{table}

%% file: sec/supp/adapt_add_models.tex
\subsection{Adaptation to More AR Models}
To validate our method, we compare baseline GRPO and STAGE across different models. Specifically, we report results for Janus-Pro 1B. Models are trained with the GenEval reward and with the HPS+GIT+Grounding DINO reward; evaluation results on GenEval, T2I-CompBench are given in Tables~\ref{table_supp_other_model_geneval}$\sim$\ref{table_supp_other_model_t2i_compbench};
Note that all training hyperparameters still follow those of Janus-Pro 7B, which may induce potential distribution collapse and result in lower metrics after RL.
In addition, we report the HPS metric: standard GRPO improves the HPSv2.1 score from 26.55 to 29.02, while our method further raises it to 29.27.

\begin{table}[h]
\centering
\setlength{\abovecaptionskip}{0.1cm}
\setlength{\belowcaptionskip}{0.1cm}
\caption{
During GRPO training the baseline collapses due to excessively large losses, degrading all metrics; adding an entropy reward prevents collapse and improves Geneval metrics. Incorporating dynamic advantage estimation and KL weighting further boosts Geneval performance, particularly on multi-object scenes.
}
\setlength{\tabcolsep}{3mm}{
\resizebox{1\columnwidth}{!}{%
\begin{tabular}{@{}lccccccc@{}}
\toprule
\multicolumn{1}{l}{Model} & Overall$\uparrow$ & Single Obj.$\uparrow$ & Two Obj.$\uparrow$ & Counting$\uparrow$ & Colors$\uparrow$ & Position$\uparrow$ & Attr. Binding$\uparrow$ \\ \midrule
Janus-Pro 1B& 0.64    & 0.96        &  0.68    & 0.41     & 0.81   & 0.45     & 0.50          \\
Baseline                  & 0.57  & 0.88 & 0.67 & 0.26 & 0.76 & 0.44 & 0.39          \\
+Ent                  & 0.78  & 0.96 & 0.91 & 0.64 & 0.85 & 0.73 & 0.62          \\
Ours                      & 0.80    & 0.93  & 0.90     & 0.66     & 0.86   & 0.73     & 0.70          \\ \bottomrule
\end{tabular}
}}
\label{table_supp_other_model_geneval}
\vspace{-3mm}
\end{table}

\begin{table}[h]
\centering
\setlength{\abovecaptionskip}{0.1cm}
\setlength{\belowcaptionskip}{0.1cm}
\caption{
Following the T2I-R1 setup, we evaluate the impact of RL on other models with T2I-compbench. Our method shows clear superiority over baseline GRPO, particularly on benchmarks involving spatial reasoning and counting.
}
\setlength{\tabcolsep}{2.5mm}{
\resizebox{1.0\columnwidth}{!}{%
\begin{tabular}{@{}lcccccccc@{}}
\specialrule{1.2pt}{0pt}{0pt} 
\multicolumn{1}{l}{Model} & Color$\uparrow$ & Shape$\uparrow$ & Texture$\uparrow$ & 2D-Spat.$\uparrow$ & 3D-Spat.$\uparrow$ & Non-Spat.$\uparrow$ & Numeracy$\uparrow$ & Complex$\uparrow$ \\ 
\midrule
Janus-Pro 1B & 0.3505   & 0.2301  & 0.2817  & 0.1073  & 0.1916  & 0.2819  & 0.2145  & 0.2730  \\
Baseline               & 0.7883 & 0.5598 & 0.7131 & 0.3495 & 0.3923 & 0.3130 & 0.5468 & 0.3844 \\
Ours                   & 0.7863 & 0.5629 & 0.7142 & 0.3637 & 0.3906 & 0.3129 & 0.5663 & 0.3860
 \\ 
\specialrule{1.2pt}{0pt}{0pt} 
\end{tabular}
}}
\label{table_supp_other_model_t2i_compbench}
\vspace{-3mm}
\end{table}

%% file: sec/supp/relation_entropy_image.tex
\section{Additional Analysis}
\subsection{Additional Analysis of Policy Entropy}
\label{supp_add_analysis_entropy}
\noindent\textbf{Relation between policy entropy \& generated Image}
For models pretrained on large-scale datasets, autoregressive (AR) training learns the data distribution. The model does not regress specific features directly; instead, it outputs a probability distribution over token indices, from which specific features are sampled using a prescribed sampling algorithm. The concentration of this distribution directly affects image quality, including content richness and accuracy.
Intuitively, higher policy entropy implies a larger exploration space and greater randomness, which tends to yield richer content, whereas lower policy entropy leads to a more precise generation process, producing images with more stable details.

To demonstrate this, we vary the sampling temperature to control entropy: higher temperature yields higher entropy, and lower temperature yields lower entropy; see Fig.~\ref{fig_supp_vis_temperature} for qualitative examples. We also measure how various metrics change with temperature (Fig.~\ref{fig_supp_score_temperature}). Larger temperatures markedly reduce structural stability, degrading metrics such as GenEval, while overly small temperatures harm diversity, reducing image-quality metrics such as HPS and ImageReward.

\input{figs/supp/supp_vis_temperature}
\input{figs/supp/supp_score_temperature}

\noindent\textbf{Policy entropy during RL}
From the experiments in Fig.~\ref{fig_supp_score_temperature}, for the chosen base model, a more peaked probability distribution benefits both the aesthetic quality of generated images and their text-following ability. Therefore, under the T2I-r1 reward configuration (HPS + Grounding DINO + GIT) and under the OCR reward, entropy tends to decrease steadily (See Fig.~\ref{fig_supp_kl_entropy_reward}), yielding a more precise distribution that improves the corresponding generation metrics.
In addition, because RL is on-policy and trains the policy model using images generated by the same policy model, the generation policy gradually becomes conservative, and the entropy itself tends to decrease.

For the GenEval reward, as discussed in the main text, its rule-based and discrete scoring mechanism means that, for many prompts, reshaping the probability distribution yields little change in the GenEval reward. As a result, policy entropy fluctuates during RL, making the original model distribution more prone to distortion and limiting RL efficiency (See Fig.~\ref{fig_supp_kl_entropy_reward}). A KL loss can mitigate this to some extent, but an overly strong KL loss can restrict model performance and encourage reward hacking. By contrast, the proposed entropy reward effectively alleviates this issue.

\input{figs/supp/supp_kl_entropy_reward}

\subsection{Additional Analysis of Entropy Regularization}
\label{supp_add_analysis_entropy_reward}

\noindent\textbf{Entropy reward or entropy loss.}
Alternatively, entropy may be incorporated directly into the loss---for example, as a $D_{\mathrm{KL}}$-style term similar to Eq.~\ref{eq_grpo_opt_objective} in mainpaper. However, this strategy is difficult to tune and often induces instability during training.
Specifically, following Eq.~\ref{eq_cal_entropy_reward} in the main text, we construct an additional loss term, denoted as $D_{\mathrm{ent}}$:

\begin{equation}
\Delta\mathcal{H}_i \;=\; \mathcal{H}_{\mathrm{ref}}(o_i) - \mathcal{H}_\theta(o_i),
\end{equation}
\begin{equation}
\mathcal{L}_{\mathrm{ent}}
\;=\; \lambda \cdot \frac{1}{G}\sum_{i=1}^G \bigl(\Delta\mathcal{H}_i\bigr)^2.
\label{eq_supp_entropy_loss}
\end{equation}

\begin{figure*}[h]
\setlength{\abovecaptionskip}{0.1cm}
\setlength{\belowcaptionskip}{0.1cm}
\begin{center}
\includegraphics[width=0.7\textwidth]{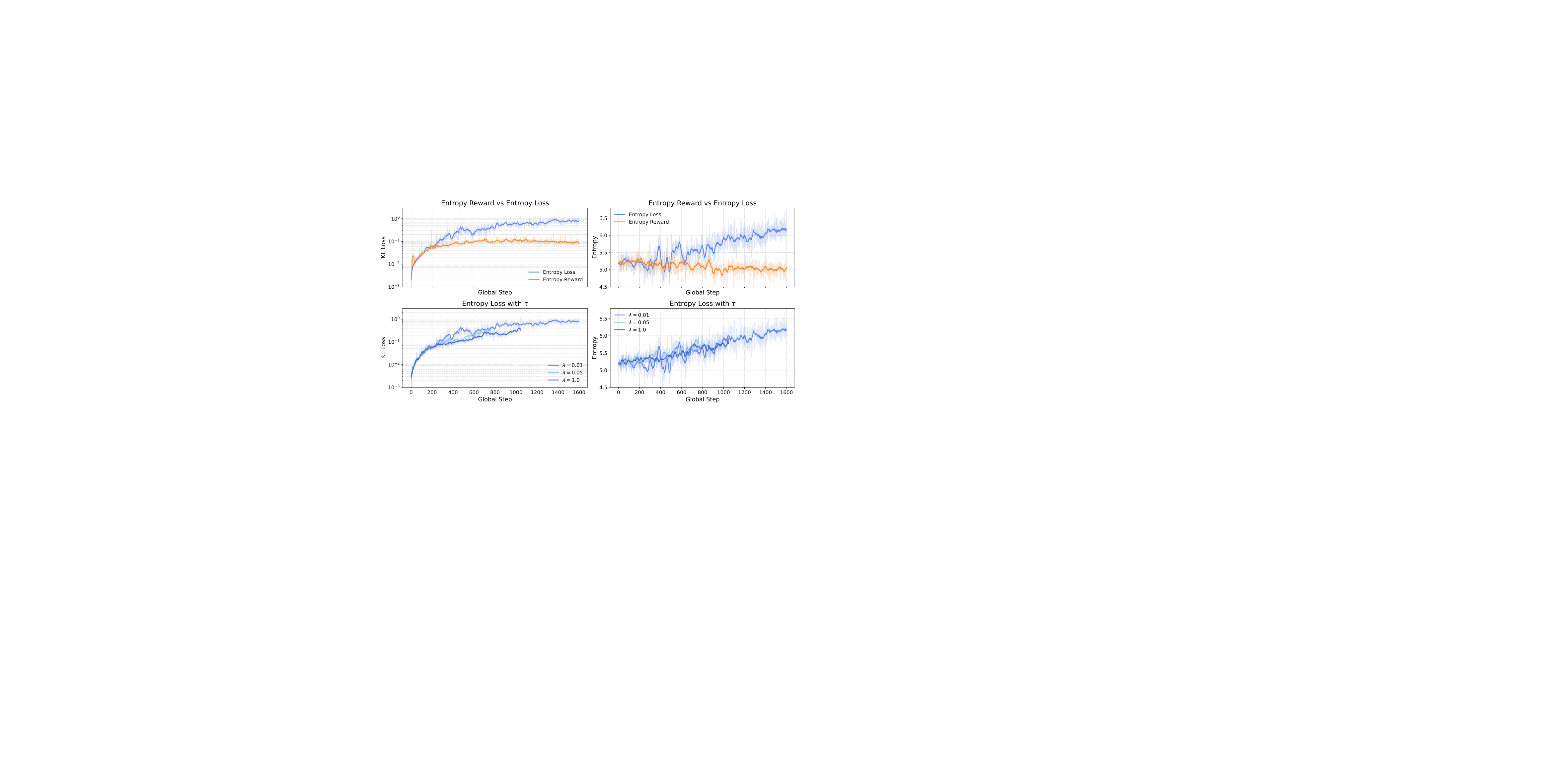}
\end{center}
\caption{
Entropy regularization applied as a loss term (coefficient $\lambda$) leads to late-stage entropy increase and consistently higher KL loss compared to our entropy-reward scheme.
}
\vspace{-0.3cm}
\label{fig_supp_entropy_reward_vs_loss}
\end{figure*}

Here $\lambda$ denotes the coefficient. We find that implementing entropy regularization as a loss term is difficult to constrain: regardless of how small $\lambda$ is set, entropy tends to increase in late-stage training (see Fig.~\ref{fig_supp_entropy_reward_vs_loss}), and the KL loss throughout training remains higher than when using our entropy-reward scheme. This may be because the loss term applies the entropy penalty uniformly to all samples, which can shift optimization away from the highest-reward samples and thus reduce the controllability of the RL process.

\begin{figure*}[ht]
\centering
\begin{minipage}[t]{0.69\textwidth}
  \centering
  \includegraphics[width=\linewidth]{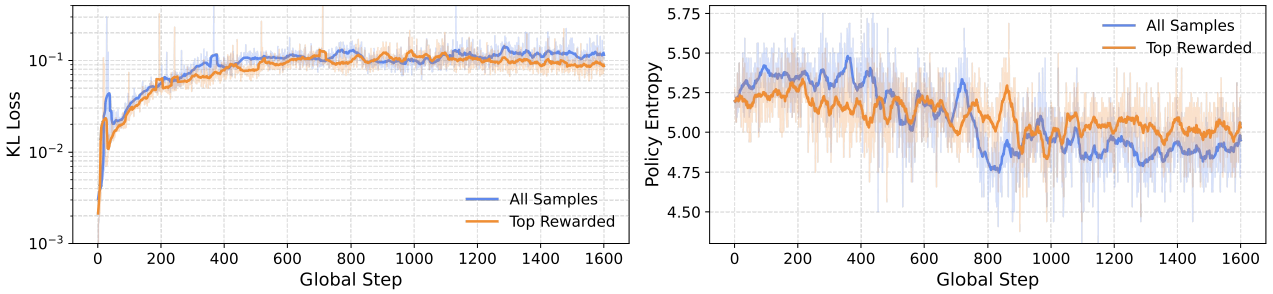}
  \captionof{figure}{
    Compared to applying the entropy loss only to the top-rewarded samples (``Top Rewarded"), applying it to all samples (``All Samples") instead tends to cause instability during training.
  }
  \label{fig_supp_kl_curves}
\end{minipage}%
\hfill
\begin{minipage}[t]{0.28\textwidth}
  \centering
  \includegraphics[width=\linewidth]{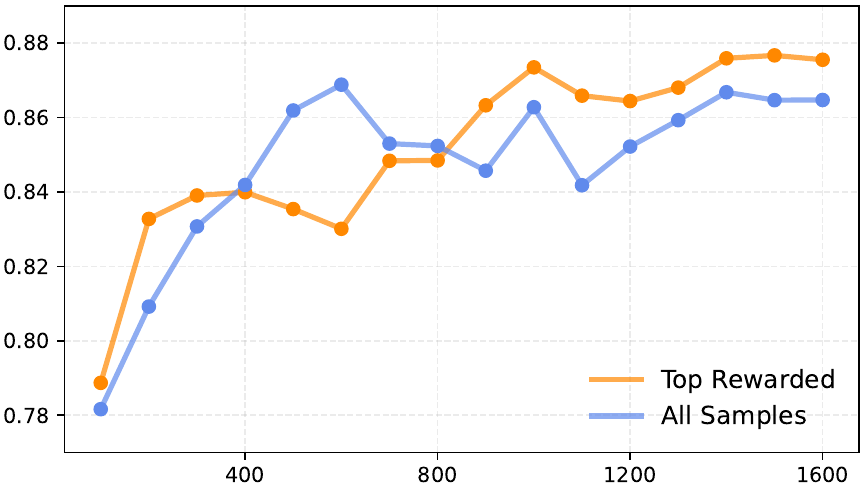}
  \captionof{figure}{GenEval during training with entropy reward on top vs. all samples.}
    \label{fig_supp_kl_curves_geneval}
\end{minipage}
\end{figure*}

\noindent\textbf{Why only reward at top-rewarded samples.}
As noted in Eq.~\ref{eq_cal_total_reward}, we apply the entropy reward only to the top-rewarded samples because uniformly adding it can bias optimization (e.g., samples with low GenEval but high entropy reward). In experiments where the entropy reward was applied to all samples, training became less stable: entropy and KL loss fluctuated more and GenEval performance deteriorated (see Fig.~\ref{fig_supp_kl_curves} and Fig.~\ref{fig_supp_kl_curves_geneval}).

%% file: figs/supp/supp_vis_temperature.tex
\begin{figure*}[h]
\setlength{\abovecaptionskip}{0.1cm}
\setlength{\belowcaptionskip}{0.1cm}
\begin{center}
\includegraphics[width=1\textwidth]{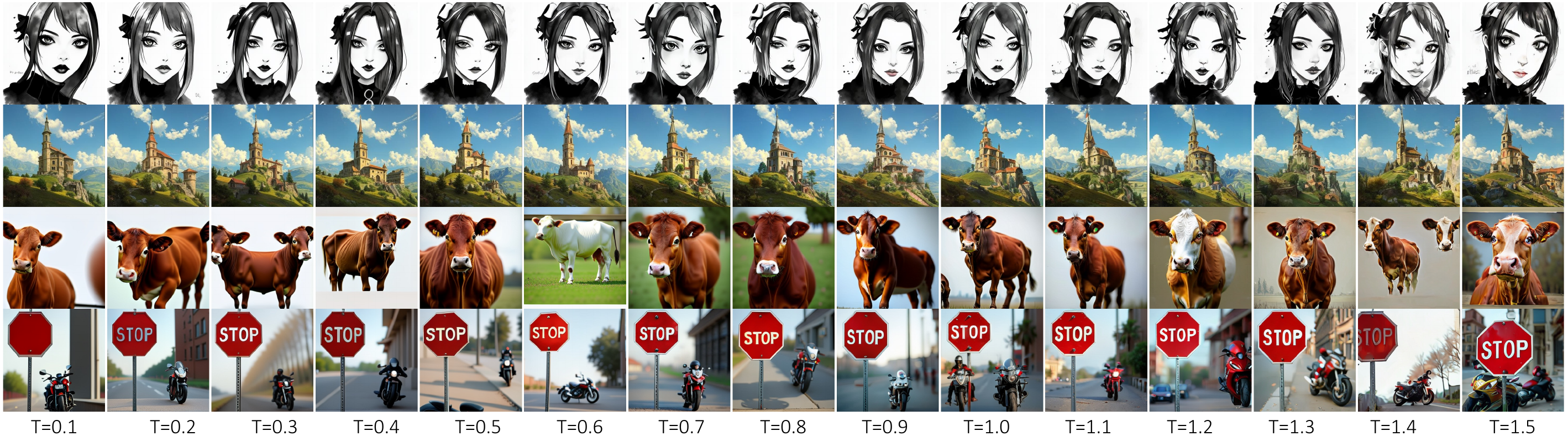}
\end{center}
\caption{
Relationship between image quality and the AR transformer’s probability distribution: lower temperature yields a more concentrated distribution and more accurate generations (especially for longer prompts such as HPS prompts; see the top two rows). An overly conservative sampling policy can still degrade image quality (e.g., GenEval prompts; see the bottom two rows). Temperatures that are too high reduce performance across prompt types.
}
\vspace{-0.3cm}
\label{fig_supp_vis_temperature}
\end{figure*}

%% file: figs/supp/supp_score_temperature.tex
\begin{figure*}[h]
\setlength{\abovecaptionskip}{0.1cm}
\setlength{\belowcaptionskip}{0.1cm}
\begin{center}
\includegraphics[width=1\textwidth]{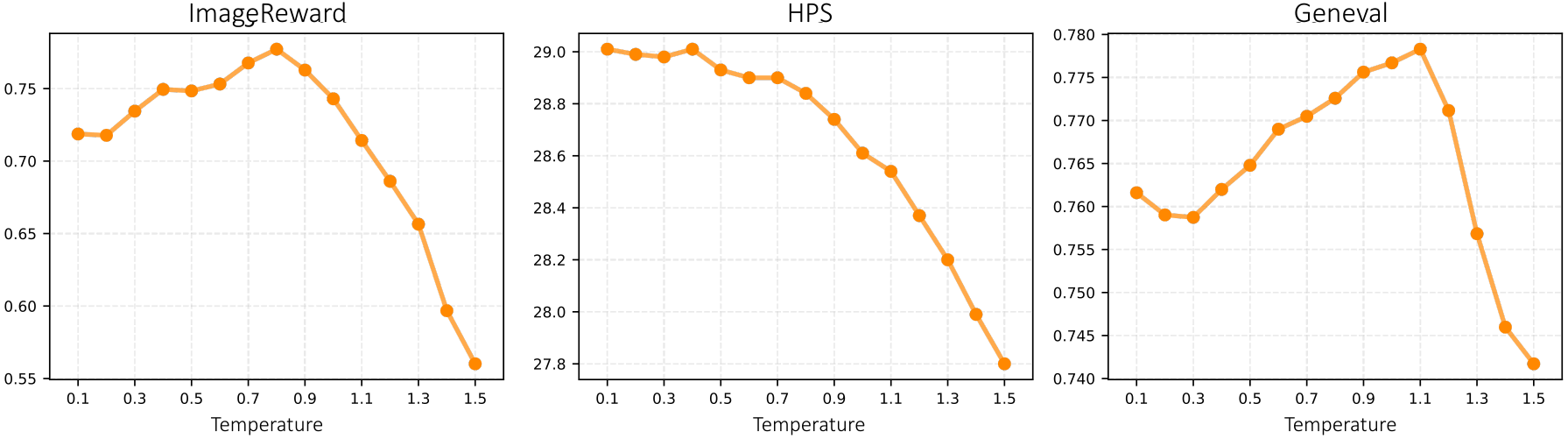}
\end{center}
\caption{
Relationship between evaluation metrics and the shape of the AR transformer’s probability distribution: adjusting the sampling temperature controls the distribution’s shape. Lower temperature makes the distribution more concentrated and the sampling policy more conservative, whereas higher temperature flattens the distribution and increases exploration.
}
\vspace{-0.3cm}
\label{fig_supp_score_temperature}
\end{figure*}

%% file: figs/supp/supp_kl_entropy_reward.tex
\begin{figure*}[t]
\setlength{\abovecaptionskip}{0.1cm}
\setlength{\belowcaptionskip}{0.1cm}
\begin{center}
\includegraphics[width=1\textwidth]{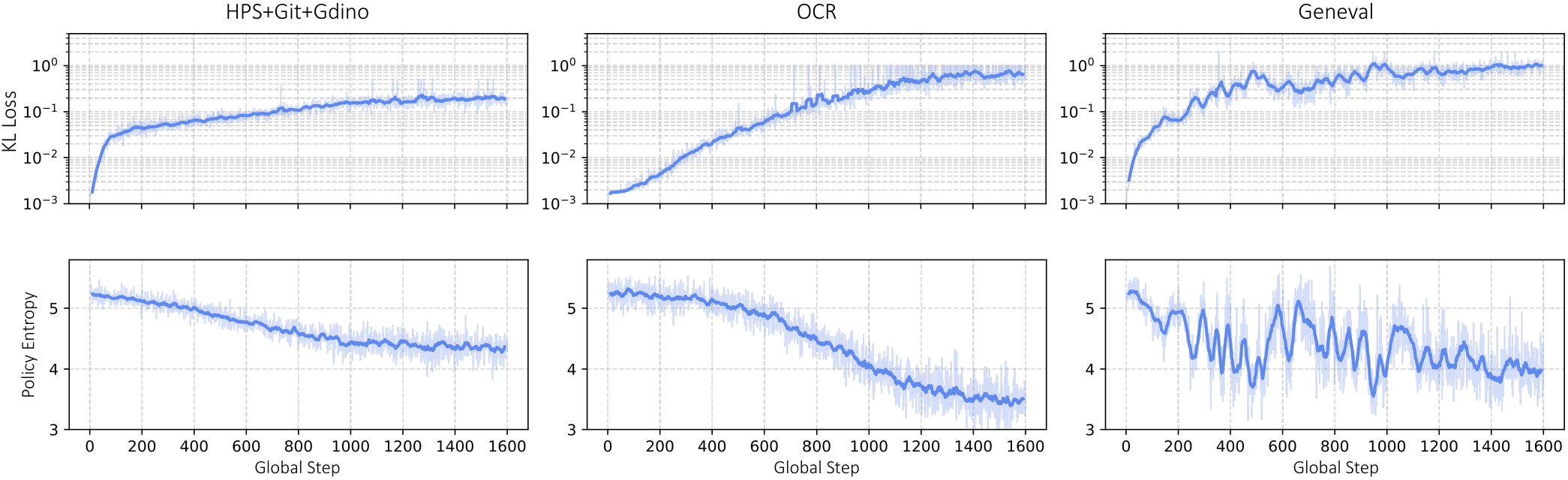}
\end{center}
\caption{
During training across different reward types, KL loss and policy entropy evolve with global step. For rewards with relatively continuous scoring (HPS + GIT + Grounding DINO, OCR), both curves change more smoothly; in contrast, with the GenEval reward, policy entropy and KL loss exhibit noticeable fluctuations.
}
\vspace{-0.3cm}
\label{fig_supp_kl_entropy_reward}
\end{figure*}

%% file: sec/supp/analysis_diverse.tex
\subsection{Additional Analysis of Image Diversity}
\label{supp_add_analysis_diversity}
The proposed similarity-based weighting method computes the dynamic weight by measuring the average similarity between positive and negative samples generated from the same prompt rollouts. This naturally raises a concern: weighting by similarity could potentially reduce diversity. To investigate this, we further analyzed the samples produced by the policy model after RL. Interestingly, we found that in some scenarios, similarity weighting may even prevent similarity collapse during RL (see Fig.~\ref{fig_supp_vis_diversity}).

This is because the RL process in AR image generation inherently tends to stabilize the sampling policy and reduce diversity, especially in the later training stages when the original model distribution is significantly distorted. Our proposed method, however, can partly preserve the original probability distribution, enabling the model to learn from reward signals while maintaining more of the initial distribution’s diversity. In contrast, the baseline tends to deviate more strongly from the original distribution in order to fit the reward model’s preferences.

Nevertheless, diversity is still sensitive to training hyper-parameters such as learning rate and KL regularization strength. In long-term training, our method may also lead to diversity reduction. To mitigate this, we recommend performing RL fine-tuning of AR models within shorter training horizons to avoid large distribution shifts caused by prolonged optimization.

\input{figs/supp/supp_vis_diversity}

%% file: figs/supp/supp_vis_diversity.tex
\begin{figure*}[h]
\setlength{\abovecaptionskip}{0.1cm}
\setlength{\belowcaptionskip}{0.1cm}
\begin{center}
\includegraphics[width=0.85\textwidth]{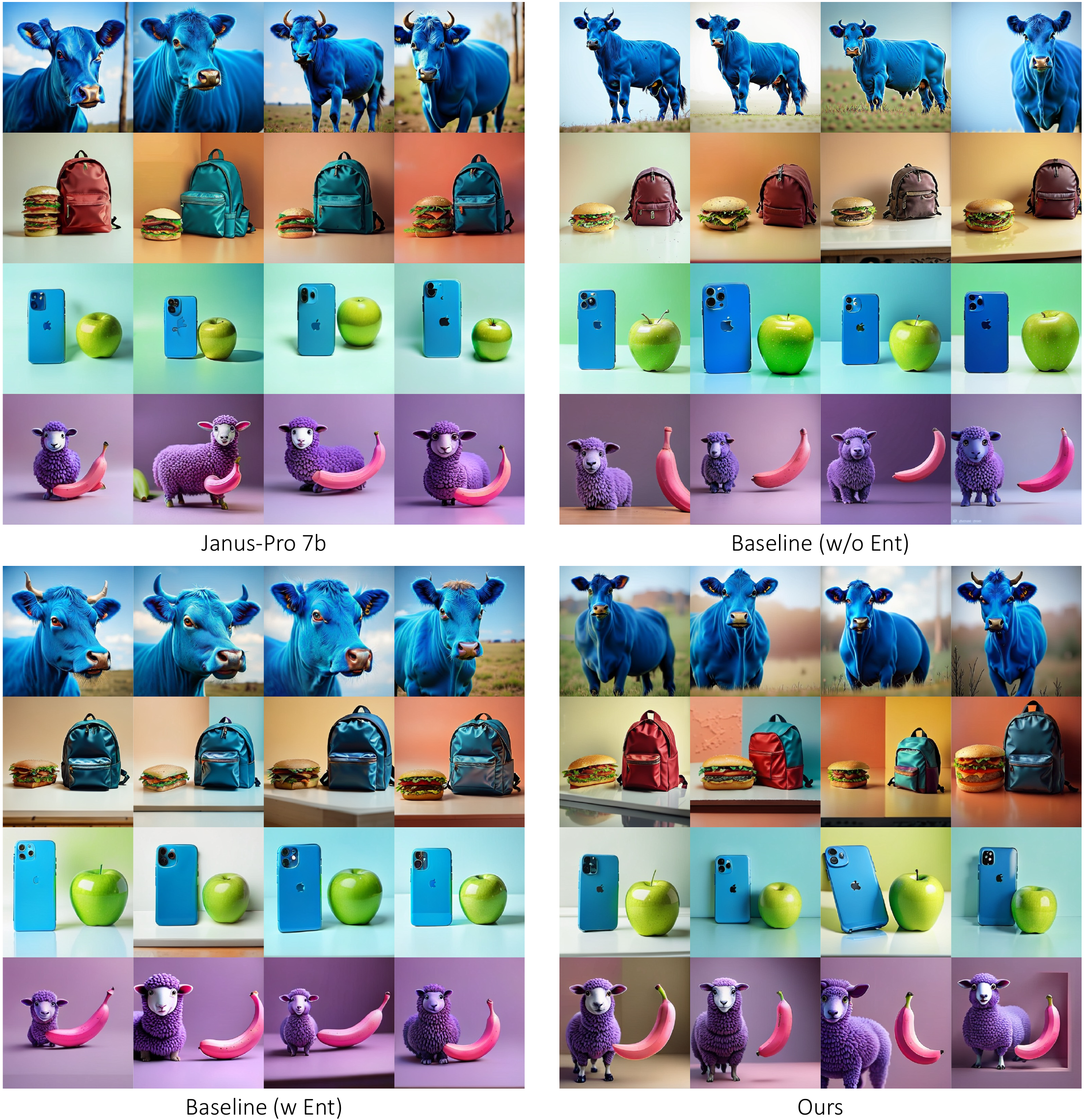}
\end{center}
\caption{
Visualization of generation diversity before and after RL. Given the same prompt, Janus-pro produces relatively diverse images, while GRPO (``w/o Ent") reduces diversity and frequently degrades image quality. Introducing an entropy reward recovers image quality to some extent (``w Ent"), and dynamic-weight RL yields the greatest variety (``Ours"). Notable differences include cow pose, backpack color, phone versus apple size, and sheep and banana angles.
}
\vspace{-0.3cm}
\label{fig_supp_vis_diversity}
\end{figure*}

%% file: sec/supp/limitations.tex
\section{Limitations $\&$ Future Works}
\subsection{Limitations}
In AR visual RL tasks, instability and reward hacking remain persistent challenges. While this work alleviates them to some extent, careful tuning of hyperparameter balances is still required to avoid degraded outcomes, and achieving long-term stable training for AR models remains an open problem.
Furthermore, our results highlight the importance of assigning different update magnitudes to individual image tokens in RL-based image generation. However, the current approach may reduce diversity, and exploring alternative token-selection or weighting strategies could further benefit training.

\subsection{Future Works}
For autoregressive (AR) generation with GRPO, although entropy reward and a KL coefficient can yield relatively stable training in practice, potential instability and image degradation risks remain. These issues stem from RL's inherent instability and from the fact that AR generation is highly sensitive to the shape of the underlying probability distribution. How RL can reliably improve AR visual generation therefore requires further investigation.
Moreover, more general problems—such as designing better rewards to balance multiple objectives and selecting training data to balance different prompt types—also merit urgent study.

%% file: sec/supp/add_vis_comp.tex
\section{Additional Visual Comparison}
\begin{figure*}[h]
\setlength{\abovecaptionskip}{0.1cm}
\setlength{\belowcaptionskip}{0.1cm}
\begin{center}
\includegraphics[width=1\textwidth]{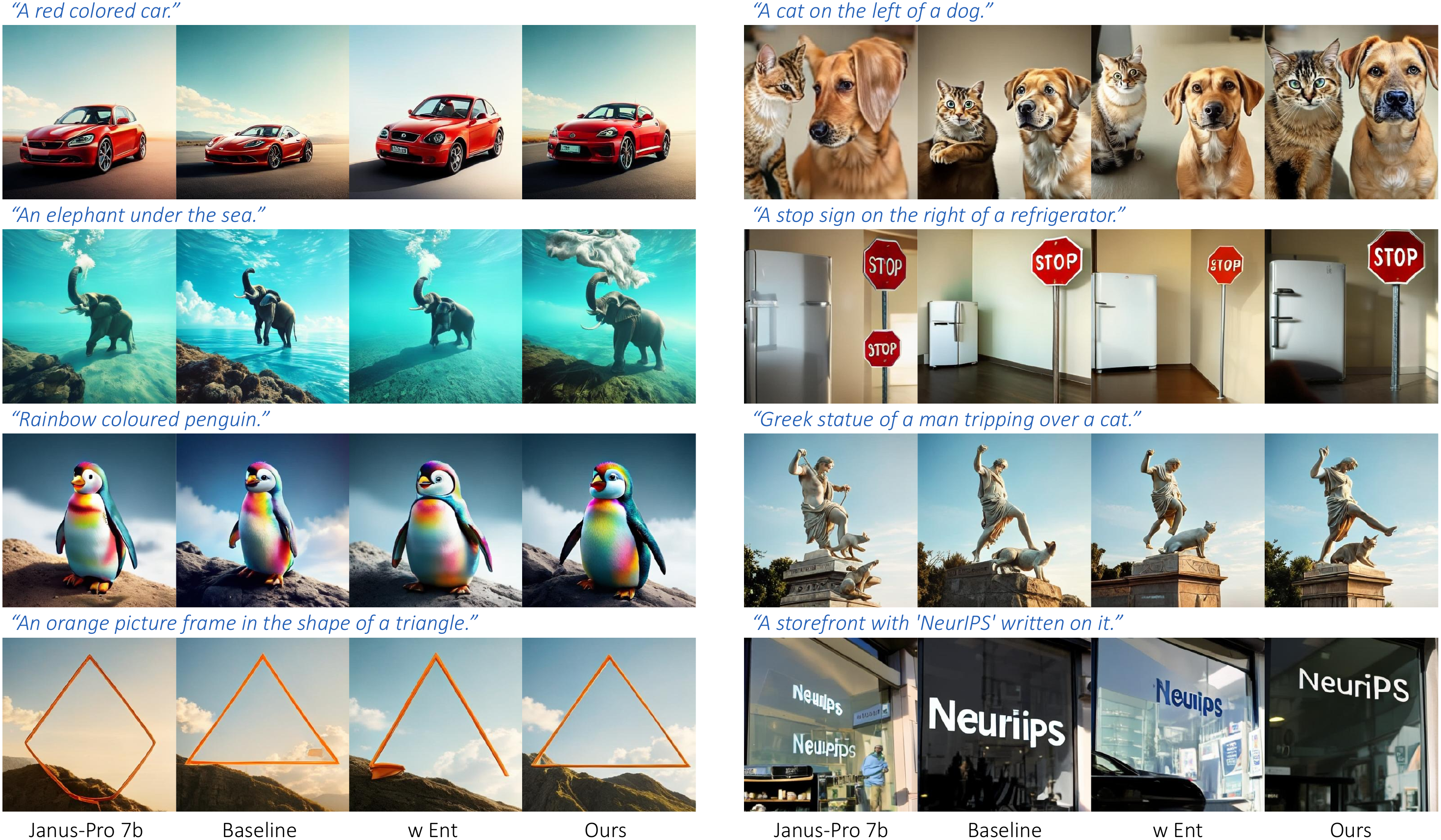}
\end{center}
\caption{
Visualization of models trained with the GenEval reward on DrawBench. By better reconciling the original model distribution with the reinforcement learning process, our method improves the fidelity and quality of generated images compared to the baseline.
}
\label{fig_supp_add_visual_comp_drawbench_genevalreward}
\end{figure*}

\begin{figure*}[h]
\setlength{\abovecaptionskip}{0.1cm}
\setlength{\belowcaptionskip}{0.1cm}
\begin{center}
\includegraphics[width=1\textwidth]{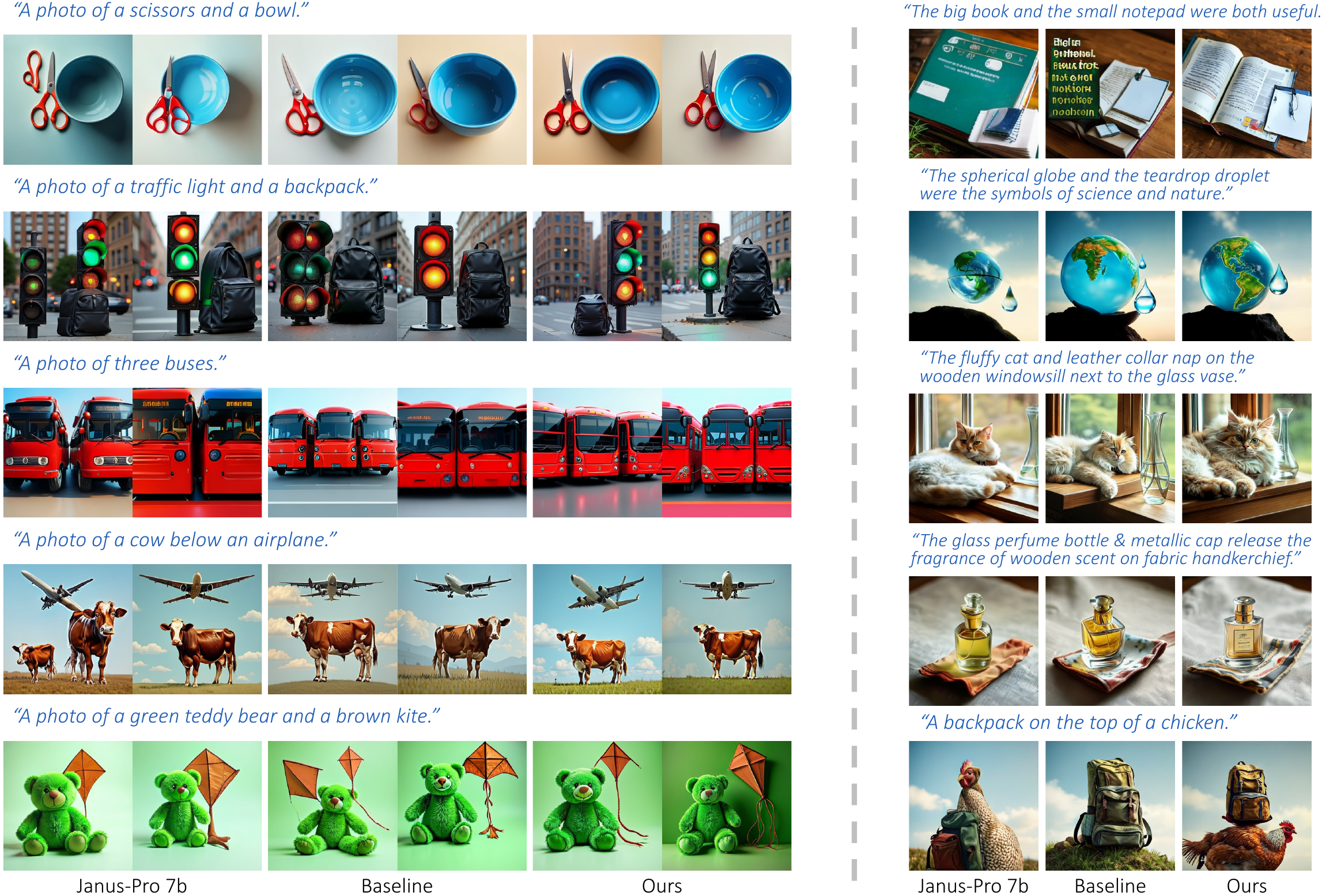}
\end{center}
\caption{
Visualization results of models trained with HPS+Gdino+Git rewards on T2I-compbench and GenEval.
}
\label{fig_supp_add_visual_comp_t2icompbench}
\end{figure*}

\begin{figure*}[h]
\setlength{\abovecaptionskip}{0.1cm}
\setlength{\belowcaptionskip}{0.1cm}
\begin{center}
\includegraphics[width=1\textwidth]{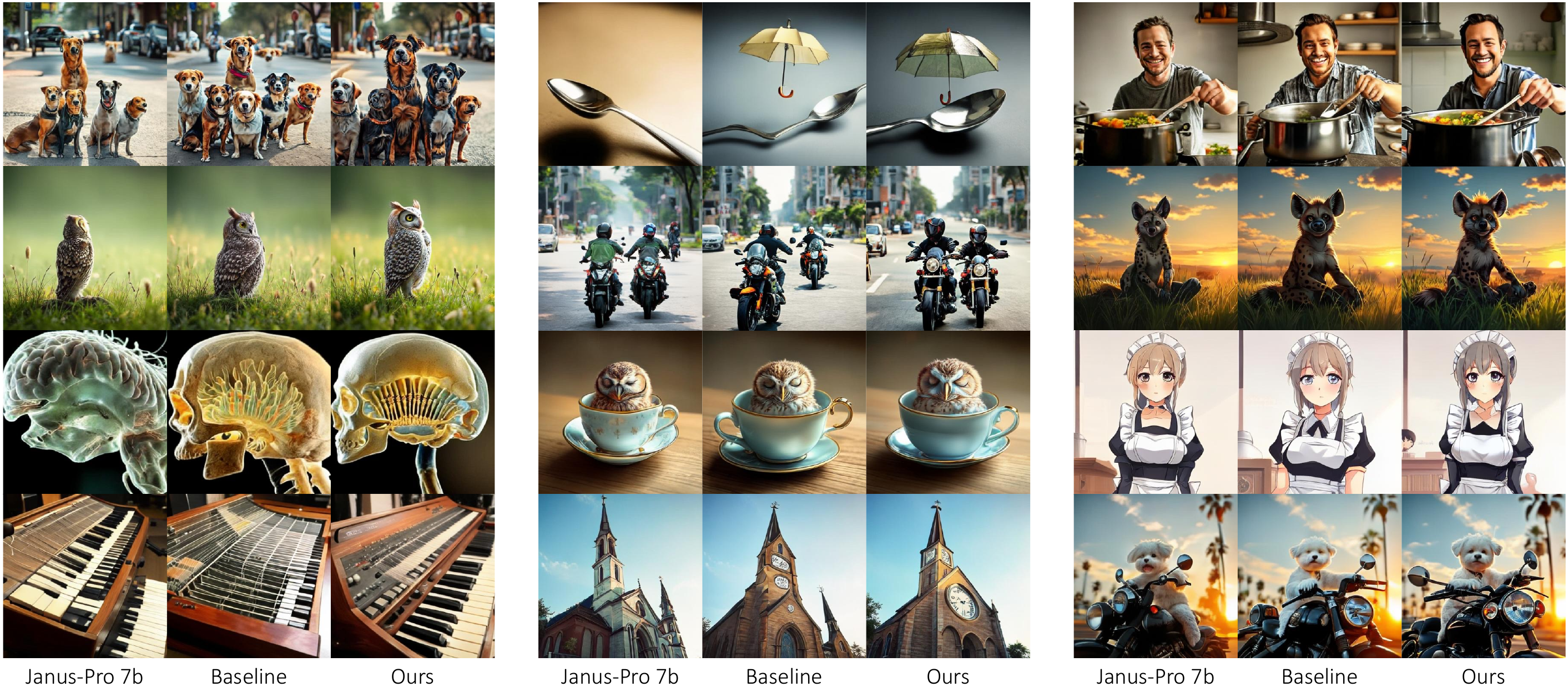}
\end{center}
\caption{
Visualization results of models trained with HPS+Gdino+Git rewards on drawbench and HPS prompts.
}
\label{fig_supp_add_visual_comp_drawbench}
\end{figure*}

We provide additional visualizations here. Fig.~\ref{fig_supp_add_visual_comp_drawbench_genevalreward} shows visual results of models trained with the GenEval reward on DrawBench (see Appendix Table~\ref{table_supp_imgrwd_geneval} for corresponding metrics). Visualizations of models trained with HPS+Gdino+Git rewards on T2I-compbench are presented in Fig.~\ref{fig_supp_add_visual_comp_t2icompbench} (see Table~\ref{tab_t2i_compbench} in the main text and Table~\ref{table_supp_geneval_generalize} in appendix). Finally, further visual examples for HPS and DrawBench are given in Fig.~\ref{fig_supp_add_visual_comp_drawbench}, with associated quantitative results reported in Table~\ref{table_hps_ocr} (main text) and Appendix Table~\ref{table_supp_imgrwd_t2i_r1}.

